\documentclass[10pt,twocolumn,letterpaper]{article}

\usepackage[pagenumbers]{cvpr} 
\usepackage{multirow}
\usepackage{booktabs}
\usepackage{array}
\usepackage{multirow}
\usepackage{rotating}
\usepackage{adjustbox}
\usepackage{xcolor}
\usepackage{algorithm}
\usepackage{algorithmic}
\usepackage{amsmath}
\usepackage{amssymb}
\usepackage{mathtools}
\usepackage{bm}

\definecolor{cvprblue}{rgb}{0.21,0.49,0.74}
\usepackage[pagebackref,breaklinks,colorlinks,allcolors=cvprblue]{hyperref}

\def\paperID{*****} 
\def\confName{CVPR}
\def\confYear{2025}

\title{NSARM: Next-Scale Autoregressive Modeling for Robust \\ Real-World Image Super-Resolution}

\author{Xiangtao Kong$^{1,2}$, Rongyuan Wu$^{1,2}$, Shuaizheng Liu$^{1,2}$, Lingchen Sun$^{1,2}$, Lei Zhang$^{1,2}$ \thanks{Corresponding author (e-mail: cslzhang@comp.polyu.edu.hk)}\\
$^{1}$The Hong Kong Polytechnic University \quad\quad\quad
$^{2}$OPPO Research Institute\\
{\tt\small \{xiangtao.kong, rongyuan.wu, Shuaizheng.liu, lingchen.sun\}@connect.polyu.hk} \\
{\tt\small {Project page: \href{https://github.com/Xiangtaokong/NSARM}{https://github.com/Xiangtaokong/NSARM}}}
}

\begin{document}

\maketitle

\begin{abstract}

Most recent real-world image super-resolution (Real-ISR) methods employ pre-trained text-to-image (T2I) diffusion models to synthesize the high-quality image either from random Gaussian noise, which yields realistic results but is slow due to iterative denoising, or directly from the input low-quality image, which is efficient but at the price of lower output quality. These approaches train ControlNet or LoRA modules while keeping the pre-trained model fixed, which often introduces over-enhanced artifacts and hallucinations, suffering from the robustness to inputs of varying degradations. Recent visual autoregressive (AR) models, such as pre-trained \textit{Infinity}, can provide strong T2I generation capabilities while offering superior efficiency by using the bitwise next-scale prediction strategy. Building upon next-scale prediction, we introduce a robust Real-ISR framework, namely Next-Scale Autoregressive Modeling (NSARM). Specifically, we train NSARM in two stages: a transformation network is first trained to map the input low-quality image to preliminary scales, followed by an end-to-end full-model fine-tuning. Such a comprehensive fine-tuning enhances the robustness of NSARM in Real-ISR tasks without compromising its generative capability. Extensive quantitative and qualitative evaluations demonstrate that as a pure AR model, NSARM achieves superior visual results over existing Real-ISR methods while maintaining a fast inference speed.  Most importantly, it demonstrates much higher robustness to the quality of input images, showing stronger generalization performance.  

\end{abstract}

\section{Introduction}
\label{sec:Introduction}

Real-world image super-resolution (Real-ISR) focuses on reconstructing high-resolution (HR) images from their low-resolution (LR) counterparts degraded by unknown real-world distortions. Unlike traditional ISR methods~\cite{dong2014learning,dong2015image,dai2019second,chen2023hat}, which focus on specific degradations and optimize MSE-based losses, Real-ISR emphasizes perceptual quality. Generative approaches such as generative adversarial networks (GANs)~\cite{goodfellow2020generative} have been widely adopted for this task. Although GAN-based methods~\cite{SRResNet,wang2018esrgan,zhang2019ranksrgan} can improve the visual perception of ISR output, they often suffer from training instability and generate many visual artifacts.

\begin{figure}[t]
\begin{center}
\includegraphics[width=1\linewidth]{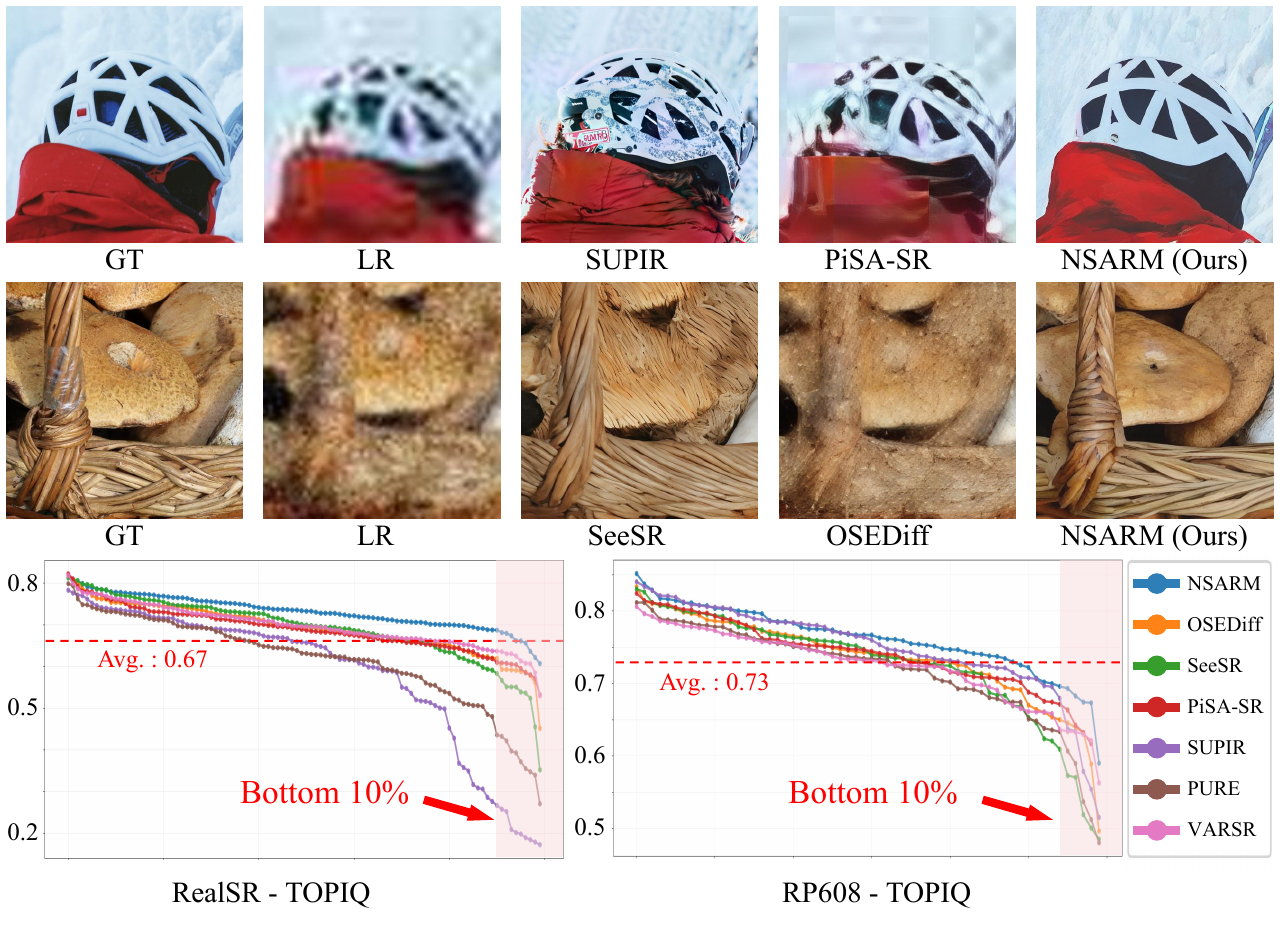}
\end{center}
\vspace{-0.3cm}
\caption{\textbf{Top two rows}: failure cases of existing Real-ISR methods, while our NSARM still works. \textbf{Bottom row}: sorted distributions of TOPIQ scores of competing methods on RealSR and RP60 datasets. We see that the quality curves of existing methods fall sharply in the late portion, indicating failure cases. For some methods, more than 10\% of the cases can fail. Our method demonstrates significantly better robustness than existing methods.}
\label{fig:intro}
\vspace{-0.5cm}
\end{figure}

The success of multimodal foundation models~\cite{CLIP,dalle1} and text-to-image (T2I) diffusion models (\eg, Stable Diffusion (SD)~\cite{sdxl,stable-diffusion3}) pretrained on vast image data has enabled significant progress in Real-ISR. Leveraging the strong generative priors of these models, many SD-based approaches have significantly surpassed GAN-based methods in visual quality. With LR images as condition, some works~\cite{wu2024seesr,yu2024scaling,yang2024pixel} synthesize HR images by iteratively denoising random Gaussian noise at higher resolution, yet they require numerous diffusion steps and suffer from slow inference. In contrast, some methods~\cite{wu2024one,sun2025pixel} directly generate HR images from LR inputs with fewer diffusion steps, greatly improving efficiency but at the cost of reduced generative capability. Meanwhile, most of the existing methods rely on ControlNet or LoRA modules to adapt a pre-trained T2I model to the Real-ISR task, often introducing over-enhanced artifacts and hallucinations, and reducing the robustness to the varying contents and degradations of the LR input (see Fig.~\ref{fig:intro} for examples). Failure cases can occur with a probability of 10\% to 20\% for some methods (see the bottom row of Fig.~\ref{fig:intro} and Fig.~\ref{fig:robust} for details). Furthermore, if we attempt to fine-tune the pre-trained SD model using task-specific losses to improve the robustness, the generative capability of the model will be degraded significantly.

Recent advances in visual autoregressive (AR) models for T2I generation may offer potential solutions to the aforementioned challenges. Instead of executing the diffusion process in high-resolution space, AR-based methods generate images by predicting visual tokens one-by-one~\cite{esser2021taming,liu2024lumina}; however, these approaches are not well suited for Real-ISR tasks. For example, PURE~\cite{wei2025perceive} applies this token-by-token generation to Real-ISR, but its iterative prediction leads to a very slow inference speed (see Tab.~\ref{tab:complexity_comparison_vertical}). A more promising remedy comes from the VAR model~\cite{VAR}. Different from SD or other AR methods, VAR introduces a next-scale prediction approach, which learns to predict residual information between different image scales, progressively refining the image from low to high resolution. This coarse-to-fine framework naturally matches the requirement of Real-ISR tasks, which model image relationships between different resolution levels. VARSR~\cite{qu2025visual} adapts VAR for Real-ISR with this paradigm. However, VARSR relies on a discrete pre-trained codebook, which fundamentally constrains the quality of reconstructed HR images. Although VARSR employs a diffusion-based post-processing module to reduce artifacts from discrete representations, this introduces additional complexity, and it is hard to obtain Real-ISR results comparable to SD-based approaches in visual quality.

\begin{figure*}[t]
\begin{center}
\includegraphics[width=1\linewidth]{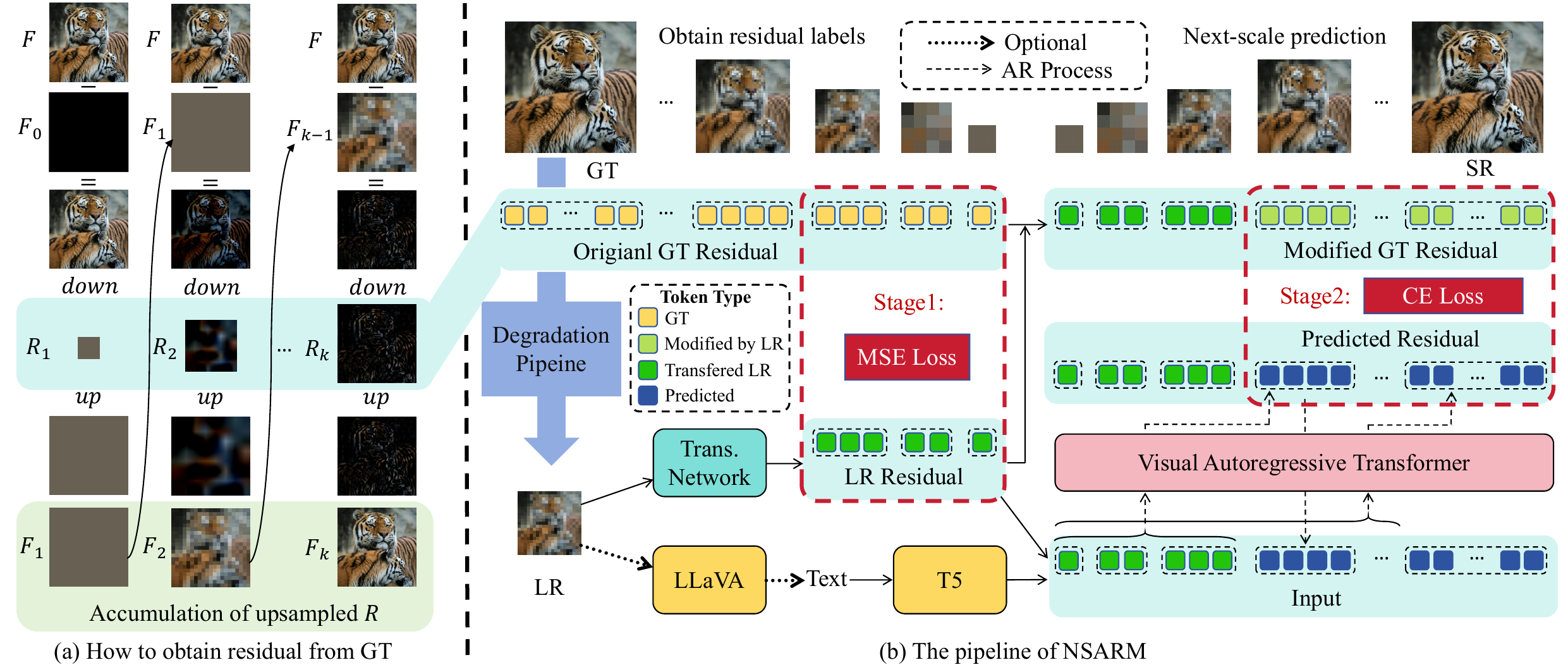}
\end{center}
\vspace{-3mm}
\caption{The image decomposition process of VAR-like methods (left) and the framework of our proposed NSARM (right). }
\label{fig:main}
\end{figure*}

To overcome the limitations of existing SD-based and AR-based Real-ISR methods, inspired by Infinity~\cite{han2024infinity}, we propose Next-Scale Autoregressive Modeling (NSARM) - a novel approach that performs next-scale prediction in a bitwise quantized space to progressively refine images from LR to HR. Via bitwise quantization, Infinity introduces the Infinite-Vocabulary Tokenizer, which provides powerful generative priors with a theoretically continuous vocabulary and high processing speed. As shown in Fig.~\ref{fig:main}, to utilize the benefits of pre-trained Infinity model, we introduce a key innovation in NSARM: a transformation network that directly processes the LR image to generate the preliminary few scales. By replacing the original scales, which could be further refined by Infinity, we could control the generation pathway. We then present a two-stage training strategy to train the NSARM model: we first train the transformation network and then fine-tune the entire network end-to-end. NSARM generates HR output from the LR image instead of random noise, and it is fully fine-tuned by the original pre-training objective and loss function; therefore, it is robust to the input LR image without sacrificing the generation capability. As can be seen from the bottom rows of Fig.~\ref{fig:intro} and Fig.~\ref{fig:robust}, NSARM shows much stronger robustness than the existing methods in Real-ISR.



Our contributions are summarized as follows. First, we propose NSARM, an efficient AR model enabled by preliminary scale transformation and bitwise next-scale prediction. Second, the proposed transformation network and two-stage training support both generation directly from LR input and full-parameter finetuning, enabling exceptional model robustness. Finally, NSARM demonstrates robust and superior visual results compared to existing SD-based methods and AR-based approaches, demonstrating outstanding potential for the Real-ISR task.

\section{Related Work}
\label{sec:Related Work}

\paragraph{Real-World Image Super-Resolution.}
Traditional ISR methods~\cite{dong2014learning,dong2015image,dai2019second,chen2023hat} typically focus on limited degradation types while optimizing with MSE-based losses, often yielding over-smoothed results with poor generalization to real-world scenarios. Driven by the growing demands for perceptual excellence in real-world deployment, recent works aim to synthesize realistic degradations to train Real-World Image Super-Resolution (Real-ISR) models. Some real-world ISR datasets have been developed \cite{realsr, drealsr}, and RealESRGAN~\cite{wang2021real} and BSRGAN~\cite{zhang2021designing} introduce  degradation pipelines that simulate real-world conditions (\eg, noise, blur, and compression artifacts), enabling single models to handle diverse degradations while improving generalization. In this work, we adopt the degradation settings of RealESRGAN to implement the model training.

\vspace{1mm}
\noindent\textbf{Diffusion-based Real-ISR Models.}
While GAN-based Real-ISR approaches~\cite{SRResNet,wang2018esrgan,zhang2019ranksrgan} can achieve improved visual quality, they suffer from training instability and limited natural image prior due to insufficient pre-training. Stable Diffusion (SD) models~\cite{sdxl,stable-diffusion3} could provide much stronger priors due to their large-scale T2I pre-training, catalyzing diverse studies. Some works directly harness these priors to boost reconstruction quality~\cite{wang2024exploiting,lin2023diffbir,yang2023pixel}, and others explore text prompting for better semantic guidance~\cite{wu2024seesr,yu2024scaling}. These methods synthesize the high-quality image from random Gaussian noise, which yields realistic results but is slow. Methods have also been developed to accelerate the Real-ISR process by reducing diffusion steps~\cite{wu2024one,yue2023resshift,sun2025pixel}, but visual quality is degraded due to the reduction in generation diversity. Most SD-based Real-ISR approaches employ generative priors by fixing the pre-trained diffusion model. However, these methods lack enough robustness and may introduce severe artifacts when handling images of varying contents and degradations.

\vspace{1mm}
\noindent\textbf{AR-based Real-ISR Models.}
Early visual autoregressive approaches like VQGAN~\cite{vqgan} and Parti~\cite{parti} employ quantization to convert images into discrete tokens for next-token prediction. PURE~\cite{wei2025perceive} introduces this paradigm to Real-ISR, but it suffers from slow inference speed. VAR~\cite{VAR} proposes next-scale prediction and significantly improves the sampling speed. Based on the VAR model, VARSR~\cite{qu2025visual} obtains fast inference speed for the Real-ISR task. However, these methods require discrete codebooks, which fundamentally limit their reconstruction capability. The Infinity~\cite{han2024infinity} model uses bitwise modeling to replace index-wise tokens with bitwise tokens, which could theoretically scale the tokenizer vocabulary to infinity. It also offers efficient token prediction, making it possible to train on a large scale. Consequently, Infinity provides more powerful generative priors with nearly continuous representations, making it well-suited for Real-ISR tasks.

\section{Method}
\label{sec:Method}

\subsection{Bitwise Visual Autoregressive Model}


\paragraph{Next-Scale Prediction.}

Traditional visual AR models first encode and decompose a GT image $\bm{I}$ to image tokens $(x_1, x_2, ..., x_T)$, then learn to predict the next token one by one from previous tokens with condition $c$:
\begin{equation}
\label{eq:ar}
    p\left(x_1, x_2, ..., x_T\right)=\prod\nolimits_{t=1}^T p\left(x_t \mid x_1, x_2, ..., x_{t-1}, c\right).
\end{equation}
As shown in Fig.~\ref{fig:main}, VAR-like models first encode the image $\bm{I}$ into latent features $\bm{F}\in \mathbb{R}^{h \times w \times d}$ and then decompose the features $\bm{F}$ into $K$ multi-scale residual maps $(\bm{R}_1, \bm{R}_2, ..., \bm{R}_K)$ by:
\begin{align}
\begin{aligned}
    \bm{R}_k =& \operatorname{down}(\bm{F} - \bm{F}_{(k-1)},(h_k,w_k)), \\
    \bm{F}_k &= \sum\nolimits_{i=1}^{k} \operatorname{up}(\bm{R}_i, (h, w)),
\end{aligned}
\label{eq:cum_sum}
\end{align}
where $\bm{F}_k$ is accumulated from $\bm{R}_1$ to $\bm{R}_k$ ($\bm{F}_0$ is zero), $\operatorname{up}(\cdot)$ and $\operatorname{down}(\cdot)$ mean upsampling and downsampling. The resolution of $\bm{R}_k$ is $h_k \times w_k$ and it grows gradually from $k=1 \to K$. Since each $\bm{R}_k$ is obtained by the GT feature $\bm{F}$ subtracting the accumulated $\bm{F}_k$, there is no information loss in the decomposition. These residual maps would be used as the input and target during training. (See details of decomposing from Alg.~\ref{alg:algorithm1} in the supplementary materials.) Then, VAR models learn to predict residuals $\bm{R}_k$ of the next scale conditioned on previous ones $(\bm{R}_1,…,\bm{R}_{k-1})$ and the condition $c$. The next-scale prediction can be formulated as: 
\begin{equation}
p(\bm{R}_1,...,\bm{R}_K)=\prod\nolimits_{k=1}^{K} p(\bm{R}_k \mid \bm{R}_1,...,\bm{R}_{k-1},c).
  \label{eq:next_scale_predict}
\end{equation}

\noindent\textbf{Infinite-Vocabulary Tokenizer and Classifier.}
Infinity~\cite{han2024infinity} adopts binary spherical
quantization (BSQ) ~\cite{zhao2024image} to increase the vocabulary size to an extremely large scale, \eg, $2^{64}$. Specifically, BSQ quantizes the continuous $\bm{R}_k \in (h_k,w_k,d)$ of Eq.~\ref{eq:cum_sum} to binary. In this way, all the possible binary combinations would form a large $2^{d}$ vocabulary
so that the infinite-vocabulary tokenizer can achieve reconstruction results comparable to the continuous encoding of SD.
Consequently, Infinity learns to predict binary residual tokens (bitwise 0/1 indices) for the next scale through cross-entropy optimization. For a $2^{d}$-sized vocabulary, the infinite-vocabulary classifier (IVC) replaces a single $2^{d}$-class classifier with $d$ parallel binary classifiers, reducing the parameter count from $O(2^{d})$ to $O(d)$. 


\subsection{The Pipeline of NSARM}
\label{sec:The Pipeline of NSARM}

\paragraph{Real-ISR Generation Pathway Control.}

During inference, Infinity generates high-resolution images through progressive scale prediction. For a target image resolution of 1024$\times$1024, there will be 13 discrete scales with resolution $\{16,32,64,96,128,192,256,320,384,512,640,768,1024\}$. Through systematic experimentation, we make two important observations about the generation process. 

\begin{figure}[t]
\begin{center}
\includegraphics[width=1\linewidth]{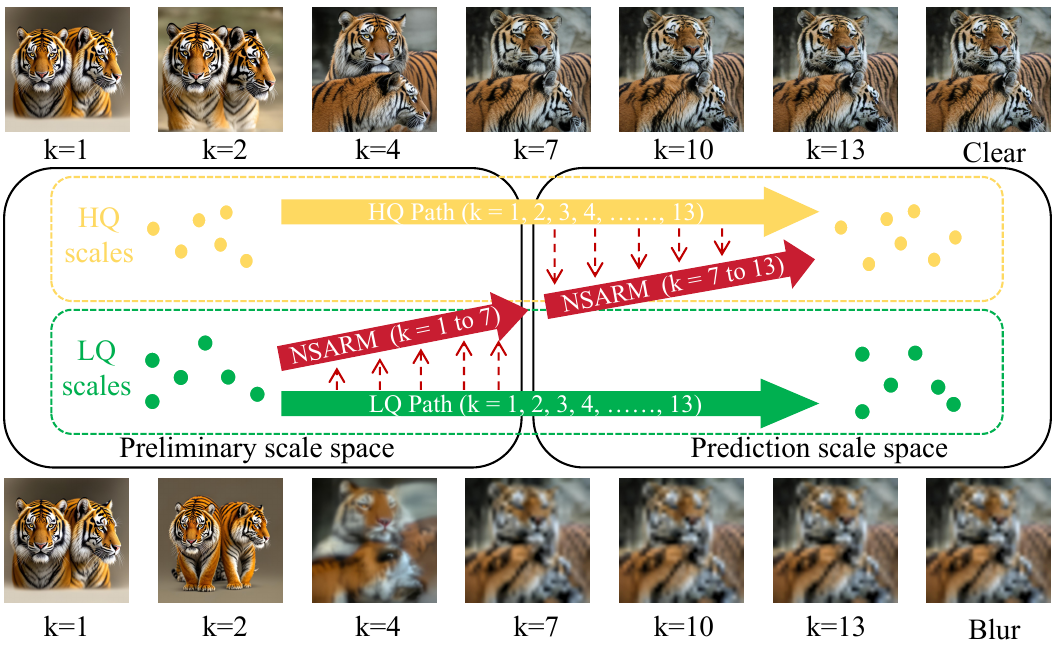}
\end{center}
\vspace{-4mm}
\caption{The top and bottom images are the generation results of Infinity with the $k$ scales replaced by clear or blurred images. NSARM is to establish a pathway towards the desired HQ by preliminary scales from the LR input.}
\label{fig:obs}
\vspace{-6mm}
\end{figure}

First, as depicted in the top row of Fig.~\ref{fig:obs}, replacing the preliminary $k$ predicted scales with the corresponding decomposed residuals from a clear image (Eq.~\ref{eq:cum_sum}, with prompt: ``Two tigers, one from the front and one from the side." ), it progressively steers the output towards the clear image with the increase of $k$. In contrast, smaller $k$ values introduce greater generation randomness. Pure T2I generation tasks correspond to $k=1$, while full replacement ($k=13$) yields the original image. The LR input provides inherent information for the preliminary scales, which indicates that the next-scale prediction has the potential to directly handle the Real-ISR task. 

Second, the bottom row of Fig.~\ref{fig:obs} demonstrates that even replacing only the earlier scales (\eg, $k=7$) with residuals derived from a blurry input, the output remains blurry, despite the model's powerful generative prior. Similarly, if preliminary scales contain degradation artifacts, they would lead to amplified artifacts in the final image. This indicates that the preliminary scales establish a critical ``generation pathway" that guides subsequent synthesis.

As shown in Fig.~\ref{fig:obs}, the key insight of our NSARM is to establish a faithful pathway towards the desired high-quality output by preliminary scales from the LR input. Our method can make better use of Infinity's powerful generative prior to achieve high-quality restoration.

\vspace{+1mm}
\noindent\textbf{Transformation Network and Two-Stage Optimization.}
To establish optimal preliminary scales for Real-ISR generation pathway control, we propose a lightweight transformation network with a dedicated two-stage optimization protocol. This network maps the input LR image into the precise preliminary scale residuals required by Infinity's image generation pathway. For 4× super-resolution targeting 1024×1024 resolution (LR input: 256×256), the transformation network outputs the first seven scales of the progressive sequence (${16,32,64,96,128,192,256}$). This design ensures: (1) full retention of LR information without spatial compression and (2) delegating preliminary degradation removal at the LR level to the transformation network, enabling the autoregressive process to focus on refined detail synthesis over subsequent scales. To this end, we introduce the following two-stage optimization strategy.

\textit{Stage 1: Pathway-Alignment by Transformation Network Training.}
As illustrated in Fig.~\ref{fig:main}, we first train the transformation network independently using MSE loss in the feature domain to align the HR generation pathway from the decomposed GT (from Alg.~\ref{alg:algorithm1}):
\begin{equation}
\mathcal{L}_{s1} = \frac{1}{K_t} \sum\nolimits_{k=1}^{K_t} | \bm{R}_k - \mathcal{T}(I_{LR})_k |^2,
\end{equation}
where $K_t=7$, $\mathcal{T}(\cdot)$ denotes the transformation network, $I_{LR}$ is the LR input, and $\bm{R}_k$ are the GT residuals for scale $k$. The network output cannot be perfectly aligned with the subsequent autoregressive process. While this stage provides basic preliminary scales, experiments in Sec.~\ref{sec:abla} reveal that using only stage 1 training will introduce artifacts. 

\begin{table*}[htb]
    \centering

    \adjustbox{width=\textwidth,center}{
\begin{tabular}{l|l|cccccccccccc}
    \toprule
    \multirow{2}{*}{Datasets} & \multirow{2}{*}{Methods} & \multirow{2}{*}{PSNR↑} & \multirow{2}{*}{SSIM↑} & \multirow{2}{*}{LPIPS↓} & \multirow{2}{*}{NIQE↓} & \multicolumn{2}{c}{MANIQA} & \multicolumn{2}{c}{MUSIQ} & \multicolumn{2}{c}{CLIPIQA}  & \multicolumn{2}{c}{TOPIQ}\\
     & & & & & & Avg.↑ & Var.↓ & Avg.↑ & Var.↓ & Avg.↑ & Var.↓ & Avg.↑ & Var.↓ \\
    \midrule
         \multicolumn{14}{c}{{Simulated Dataset (Evaluated on generating 1024 $\times$ 1024 images)}} \\
    \midrule
         \multirow{5}{*}{DIV2K} 
           & SeeSR~\cite{wu2024seesr} & \textcolor{red}{22.34} & \textcolor{red}{0.5757} & 0.3452 & 4.13 & \textcolor{blue}{0.6209} & 0.0037 & \textcolor{blue}{71.72} & \textcolor{blue}{24.7735} & \textcolor{blue}{0.7074} & 0.0119 & \textcolor{red}{0.6878} & 0.0068 \\
      & SUPIR~\cite{yu2024scaling} & 21.14 & 0.5248 & 0.3784 & \textcolor{red}{3.43} & \textcolor{red}{0.6266} & 0.0084 & 70.19 & 61.8712 & 0.6096 & 0.0190 & 0.6266 & 0.0115 \\
      & OSEDiff~\cite{wu2024one} & 22.10 & \textcolor{blue}{0.5681} & \textcolor{blue}{0.3389} & 3.89 & 0.5933 & \textcolor{red}{0.0031} & 70.43 & 25.1750 & 0.6637 & 0.0126 & 0.6177 & 0.0047 \\
      & PiSA-SR~\cite{sun2025pixel} & \textcolor{blue}{22.23} & 0.5660 & \textcolor{red}{0.3380} & \textcolor{blue}{3.78} & 0.6202 & \textcolor{blue}{0.0033} & 71.70 & 32.4889 & 0.7018 & \textcolor{blue}{0.0111} & \textcolor{blue}{0.6747} & \textcolor{blue}{0.0045} \\
      & NSARM (Ours) & 20.17 & 0.4950 & 0.3644 & 3.93 & 0.6094 & 0.0036 & \textcolor{red}{73.50} & \textcolor{red}{10.1565} & \textcolor{red}{0.7206} & \textcolor{red}{0.0082} & 0.6689 & \textcolor{red}{0.0025} \\
    \midrule
         \multicolumn{14}{c}{{Real-world Datasets (Evaluated on generating 512 $\times$ 512 images)}} \\
    \midrule
         \multirow{7}{*}{DRealSR} 
      & SeeSR~\cite{wu2024seesr} & \textcolor{blue}{28.76} & 0.7762 & 0.3316 & 7.51 & 0.5596 & 0.0124 & 61.11 & 174.4791 & 0.6401 & 0.0217 & 0.6217 & 0.0207 \\
 & SUPIR~\cite{yu2024scaling} & 25.00 & 0.6373 & 0.4334 & 7.76 & 0.5530 & 0.0117 & 59.75 & 192.8875 & 0.6921 & 0.0217 & 0.5871 & 0.0296 \\
& OSEDiff~\cite{wu2024one} & 28.60 & \textcolor{blue}{0.7872} & \textcolor{blue}{0.3296} & 7.94 & 0.5373 & 0.0087 & 60.91 & 142.5503 & 0.7082 & 0.0161 & 0.6208 & 0.0113 \\
& PiSA-SR~\cite{sun2025pixel} & \textcolor{red}{28.88} & \textcolor{red}{0.7974} & \textcolor{red}{0.3145} & 7.81 & 0.5243 & 0.0091 & 60.33 & 131.9326 & 0.6453 & 0.0217 & 0.5987 & 0.0130 \\
 & PURE~\cite{wei2025perceive} & 25.10 & 0.6184 & 0.4515 & \textcolor{red}{6.46} & 0.5856 & \textcolor{blue}{0.0068} & 62.38 & 113.5270 & 0.6694 & 0.0165 & 0.5888 & 0.0108 \\
 & VARSR~\cite{qu2025visual} & 28.22 & 0.7660 & 0.3530 & \textcolor{blue}{6.89} & \textcolor{red}{0.6146} & \textcolor{red}{0.0052} & \textcolor{red}{68.18} & \textcolor{red}{81.4850} & \textcolor{blue}{0.7216} & \textcolor{red}{0.0097} & \textcolor{blue}{0.6793} & \textcolor{blue}{0.0057} \\
 & NSARM (Ours) & 26.66 & 0.6914 & 0.4038 & 7.56 & \textcolor{blue}{0.5903} & 0.0069 & \textcolor{blue}{67.67} & \textcolor{blue}{89.8052} & \textcolor{red}{0.7259} & \textcolor{blue}{0.0103} & \textcolor{red}{0.6991} & \textcolor{red}{0.0054} \\
    \midrule
         \multirow{7}{*}{RealSR} 
      & SeeSR~\cite{wu2024seesr} & 25.52 & \textcolor{blue}{0.7369} & \textcolor{blue}{0.3065} & 6.79 & 0.6202 & 0.0056 & 68.57 & 44.2618 & 0.6663 & 0.0141 & 0.6909 & 0.0063 \\
 & SUPIR~\cite{yu2024scaling} & 23.69 & 0.6632 & 0.3568 & \textcolor{blue}{6.02} & 0.5780 & 0.0099 & 61.72 & 166.3463 & 0.6594 & 0.0243 & 0.5857 & 0.0291 \\
& OSEDiff~\cite{wu2024one} & 25.52 & 0.7334 & 0.3275 & 6.66 & 0.6267 & 0.0043 & 68.65 & 34.3655 & \textcolor{red}{0.7393} & 0.0081 & \textcolor{blue}{0.6924} & 0.0035 \\
& PiSA-SR~\cite{sun2025pixel} & \textcolor{red}{25.91} & \textcolor{red}{0.7526} & \textcolor{red}{0.2902} & 6.35 & 0.6254 & \textcolor{blue}{0.0031} & 69.14 & 29.6005 & 0.6803 & 0.0104 & 0.6885 & 0.0029 \\
 & PURE~\cite{wei2025perceive} & 22.61 & 0.6029 & 0.3848 & \textcolor{red}{5.51} & 0.6280 & 0.0034 & 66.89 & 50.1819 & 0.6909 & 0.0100 & 0.6169 & 0.0116 \\
 & VARSR~\cite{qu2025visual} & \textcolor{blue}{25.55} & 0.7263 & 0.3218 & 6.05 & \textcolor{red}{0.6541} & \textcolor{red}{0.0030} & \textcolor{red}{71.30} & \textcolor{blue}{24.5466} & 0.6996 & \textcolor{blue}{0.0068} & 0.6978 & \textcolor{blue}{0.0025} \\
 & NSARM (Ours) & 23.52 & 0.6399 & 0.3706 & 6.81 & \textcolor{blue}{0.6402} & 0.0037 & \textcolor{blue}{71.09} & \textcolor{red}{24.5262} & \textcolor{blue}{0.7370} & \textcolor{red}{0.0061} & \textcolor{red}{0.7318} & \textcolor{red}{0.0016} \\
    \midrule
         \multirow{7}{*}{RP60} 
      & SeeSR~\cite{wu2024seesr} & - & - & - & 5.16 & 0.6717 & \textcolor{blue}{0.0018} & 71.86 & 21.8666 & 0.7881 & 0.0184 & 0.7253 & 0.0061 \\
 & SUPIR~\cite{yu2024scaling} & - & - & -& \textcolor{blue}{4.66} & 0.6611 & 0.0037 & 71.80 & 20.8673 & 0.8275 & 0.0096 & \textcolor{blue}{0.7490} & 0.0041 \\
  & OSEDiff~\cite{wu2024one} & - & -& -& 5.49 & 0.6589 & \textcolor{red}{0.0016} & 71.84 & 16.9560 & \textcolor{blue}{0.8413} & 0.0069 & 0.7351 & 0.0036 \\
       & PiSA-SR~\cite{sun2025pixel} & - & - & - & 5.70 & 0.6597 & 0.0021 & 71.18 & \textcolor{blue}{15.7543} & 0.8145 & 0.0090 & 0.7282 & \textcolor{blue}{0.0023} \\
   & PURE~\cite{wei2025perceive} & - & - & - & \textcolor{red}{4.57} & \textcolor{red}{0.6973} & 0.0023 & 73.31 & 21.4124 & 0.8065 & 0.0087 & 0.7164 & 0.0048 \\
      & VARSR~\cite{qu2025visual} & - & -&- & 5.26 & 0.6681 & 0.0026 & \textcolor{blue}{73.82} & 17.9560 & 0.7709 & \textcolor{blue}{0.0065} & 0.7238 & 0.0026 \\
      & NSARM (Ours) & - &-& - & 5.69 & \textcolor{blue}{0.6862} & 0.0027 & \textcolor{red}{73.95} & \textcolor{red}{11.5072} & \textcolor{red}{0.8564} & \textcolor{red}{0.0040} & \textcolor{red}{0.7615} & \textcolor{red}{0.0021} \\
    \bottomrule
    \end{tabular} 

    }
\caption{Performance comparison of different Real-ISR methods on commonly used datasets. \textcolor{red}{Red} and \textcolor{blue}{blue} colors represent the best and second best performance. Symbols ↓ and  ↑ represent that the smaller or bigger is better.}
\label{tab:methods_comparison}
\vspace{-0.2cm}
\end{table*}

\textit{Stage 2: Next-Scale Autoregressive Fine-tuning.}
As shown in Fig.~\ref{fig:main}, we then fine-tune the whole NSARM model, including the visual autoregressive transformer, given the transformed preliminary LR scales $(\bm{R}_1,…,\bm{R}_{K_t})$ from LR. The next-scale prediction of subsequent scales can be formulated as follows:
\begin{equation}
p(\bm{R}_{K_t+1},...,\bm{R}_K)=\prod\nolimits_{k=K_t+1}^{K} p(\bm{R}_k \mid \bm{R}_1,...,\bm{R}_{k-1},c),
\label{eq:next_scale_predict_finetune}
\end{equation}
where $K_t=7$, and $c$ is the description text of the LR image. Following Infinity's original pre-training, we use bitwise cross-entropy loss to supervise binary token predictions, addressing the exponentially large token space in high-resolution scale prediction:
\begin{equation}
\mathcal{L}_{s2} = -\frac{1}{N} \sum\nolimits_{i=1}^{N}  \mathbf{r}^{\text{MGT}}_i \cdot \log(p(\mathbf{r}^{\text{pred}}_i)) ,
\label{eq:ce_loss}
\end{equation}
where $N = h_k \times w_k$ denotes the total tokens on scale $k$, $\mathbf{r}^{\text{MGT}}_i \in \{0,1\}$ is the GT binary token which has been modified by transformed preliminary LR scales $(\bm{R}_1,…,\bm{R}_{K_t})$ through Eq.~\ref{eq:cum_sum} (More details are in Alg.~\ref{alg:algorithm2} of he supplementary materials.), and $p(\mathbf{r}^{\text{pred}}_i)$ is the predicted probability of token $i$. It should be noted that the fine-tuning protocol at this stage preserves Infinity's original pre-training objective and constraints. The only modification is that the preliminary scales are now supplied by the transformation network rather than being generated from the condition text. This strategic design allows NSARM to maximally retain Infinity's inherent generative priors while adapting the generation pathway to Real-ISR requirements.

Our two-stage optimization framework offers both training stability and convergence efficiency. Stage 1 provides a meaningful initialization by learning LR-to-residual mappings, while the fine-tuning in Stage 2 focuses on accurate adaptation. For a better understanding of our training process, the pseudo codes of our algorithm are summarized in the \textbf{supplementary materials}. As also demonstrated in Sec.~\ref{sec:abla}, direct Stage 2 optimization without Stage 1 initialization (analogous to conditional diffusion model training) exhibits very slow convergence.



\begin{figure*}[t]
\begin{center}
\includegraphics[width=1\linewidth]{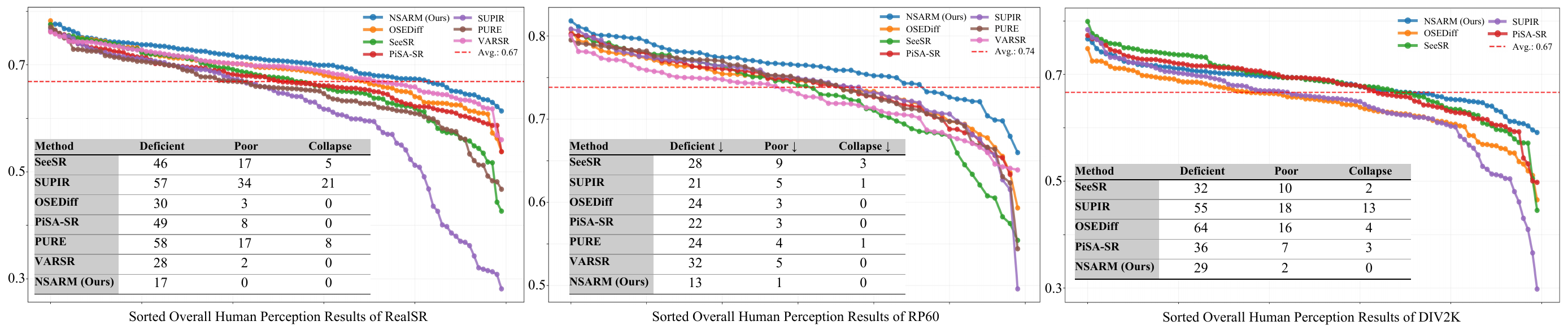}
\end{center}
\vspace{-0.3cm}
\caption{The model robustness performance. The curves show the sorted average scores over CLIPIQA, MUSIQ, MANIQA (divided by 100) and TOPIQ for different methods on the RealSR, RP60 and DIV2K datasets. The table in each sub-figure lists the numbers of Deficient, Poor and Collapse cases for each method.}
\label{fig:robust}
\vspace{-0.3cm}
\end{figure*}

\section{Experiments}
\label{sec:Experiments}

\subsection{Experimental Settings}

\paragraph{Training Details.}
We train NSARM for $\times 4$ Real-ISR with $1024 \times 1024$ target resolution using the AdamW optimizer~\cite{adamw}. In the first stage, we separately train the transformation network for 100K iterations, using a learning rate of $2 \times 10^{-5}$ and a large batch size of 196 to ensure stable convergence. The second stage fine-tunes the complete model on the pre-trained 8B Infinity \cite{han2024infinity} backbone for 40K iterations with learning rate $4\times10^{-5}$ and batch size 32. The training is implemented using 16 NVIDIA A800 GPUs. Our training data comprises 1M $1024\times1024$ high-quality images with text annotation from Unsplash~\cite{unsplash}. LR-GT pairs are synthesized using RealESRGAN's degradation pipeline.

\vspace{1mm}
\noindent\textbf{Compared Methods.}
We compare NSARM with state-of-the-art SD-based and AR-based Real-ISR methods,
including SeeSR~\cite{wu2024seesr}, OSEDiff~\cite{wu2024one}, PiSA-SR~\cite{sun2025pixel}, SUPIR~\cite{yu2024scaling}, PURE~\cite{wei2025perceive} and VARSR~\cite{qu2025visual}. Among them, SeeSR, OSEDiff and PiSA-SR are built upon SD 2.1 \cite{sd2.1} and SUPIR is built upon SDXL \cite{sdxl}, while all of them could generate $1024 \times 1024$ images.
SeeSR and SUPIR generate HR from random Gaussian noise while OSEDiff and PiSA-SR generate HR from LR inputs. PURE and VARSR are AR-based methods (VARSR also adopts diffusion-based refinement), which only support generating $512 \times 512$ images. 

\vspace{1mm}
\noindent\textbf{Testing Protocol.}
Since NSARM is trained for $1024\times1024$ resolution while some methods only support $512\times512$ resolution generation, and the commonly used real-world test images are also of resolution $512\times512$, we conduct evaluations at both resolutions to ensure a fair comparison. For $1024\times1024$ resolution, we collect 100 images cropped from DIV2K Validation~\cite{div2k} and degrade them using the RealESRGAN degradation pipeline to synthesize the LR input. For $512\times512$ resolution, following previous works, we adopt the RealSR~\cite{realsr} and DrealSR~\cite{drealsr} datasets, which contain LR-GT pairs, and the RP60~\cite{yu2024scaling} dataset, which only has LR images without GT. For methods that can generate $1024\times1024$ resolution images, we first generate the HR images and then downsample (using bicubic downsampling) them to $512\times512$ for comparison. For methods SeeSR, OSEDiff and PiSA-SR, which can generate $1024\times1024$ resolution images but are trained on $512\times512$ resolution, we also provide their direct $512\times512$ Real-ISR results in the \textbf{supplementary materials}.

\vspace{1mm}
\noindent\textbf{Evaluation Protocol.} We comprehensively evaluate the competing methods from different aspects, including \textit{output quality}, \textit{model complexity}, and \textit{model robustness}.

First, following previous work, we assess the model performance using full-reference metrics (PSNR and SSIM computed on the YCbCr space's Y channel; LPIPS~\cite{lpips} in RGB space) and no-reference metrics (NIQE~\cite{niqe}, CLIPIQA~\cite{clipiqa}, MUSIQ~\cite{musiq}, MANIQA~\cite{maniqa}, and TOPIQ~\cite{chen2024topiq}). We also provide qualitative comparisons to validate the visual effects of the proposed NSARM.
Second, we evaluate the model complexity in terms of the number of parameters and inference time. 

Third and more importantly, we evaluate the model robustness by using four human perception metrics (CLIPIQA, MUSIQ, MANIQA and TOPIQ), which can reflect the stability of models in practical use. We measure the robustness from two aspects. The first is the variance of the perception metrics over each test dataset, where a smaller variance indicates a more robust performance. Second, we sort the perception metric scores from high to low, and count the number of failure cases (\ie, long-tail cases) to evaluate the model robustness. Specifically, we calculate the global mean, denoted by $\mu_G$, for a metric on a dataset, then count the number of failure cases under three different levels:
\begin{align}
\begin{aligned}
    \text{Deficient} & : \sum\nolimits_{i=1}^N 1 {[x_i <  \mu_G]}, \\
    \text{Poor} & :  \sum\nolimits_{i=1}^N 1 {[x_i < 0.9 * \mu_G]}, \\
    \text{Collapse} & : \sum\nolimits_{i=1}^N 1 {[x_i < 0.8 * \mu_G]},
\end{aligned}
\label{eq:robustness}
\end{align}

where Deficient, Poor and Collapse mean the relative image quality. Clearly, the model with better robustness should have fewer failure cases.

\begin{figure*}[t]
\begin{center}
\includegraphics[width=1\linewidth]{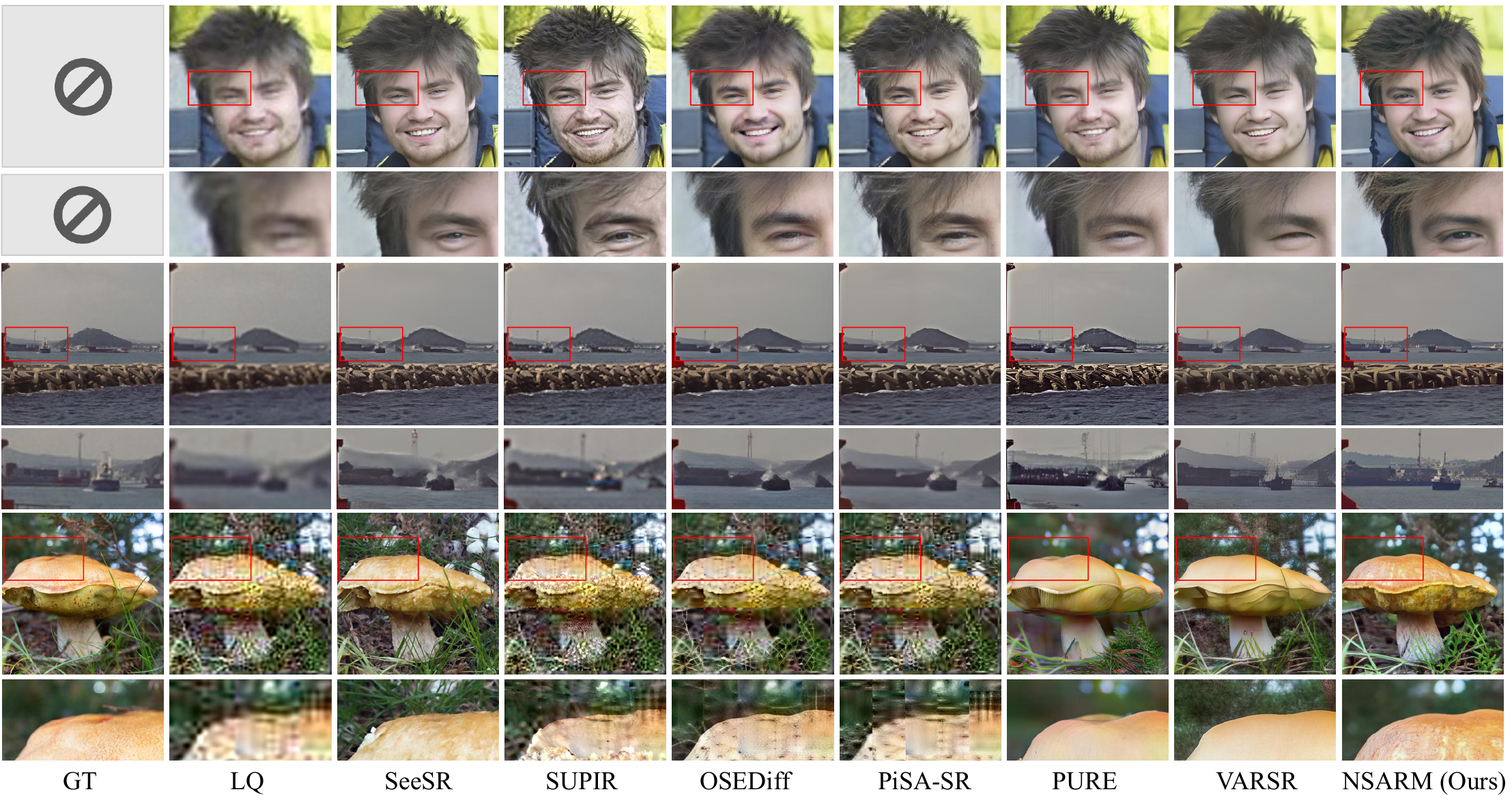}
\end{center}
\vspace{-5mm}
\caption{Visual comparison of different methods on RP60 (no GT), RealSR and DIV2K datasets (zoom in for a better view).}
\label{fig:visual}
\vspace{-0.2cm}
\end{figure*}

\subsection{Comparison with State-of-the-Arts}

\paragraph{Quantitative Quality.}

Tab.~\ref{tab:methods_comparison} compares our NSARM with other SD-based and AR-based methods. We can have the following observations. First, for full-reference fidelity metrics (PSNR/SSIM/LPIPS), those SD-based methods (PiSA-SR, OSEDiff and SeeSR) and AR-SD hybrid method (VARSR) are advantageous because they employ a continuous VAE to represent the image, naturally leading to better reconstruction fidelity. Nonetheless, NSARM achieves better fidelity metrics than another pure AR method - PURE \cite{wei2025perceive}. Second, for no-reference perceptual metrics (CLIPIQA, MUSIQ, MANIQA and TOPIQ), which can better measure the user's subjective experience, NSARM obtains overall better results than its competitors, because it tends to generate richer, more natural details favored by human perception. VARSR also shows strong performance in no-reference metrics. However, it integrates the AR-based VAR model and a diffusion-based refinement, making the whole Real-ISR process rather complex. As we will see in the qualitative comparison, VARSR actually tends to generate smooth images.

\vspace{1mm}
\noindent\textbf{Robustness.}
Beyond the overall best results in human perception metrics, our NSARM demonstrates strong robustness to input images. First, as can be seen in  Tab.~\ref{tab:methods_comparison}, NSARM achieves the lowest variances of many perception metrics, indicating that it could perform stably for LR inputs of varying contents and degradations. 

To more clearly demonstrate the robustness of NSARM, in Fig.~\ref{fig:robust} we plot the curves of sorted average scores over CLIPIQA, MUSIQ, MANIQA (divided by 100) and TOPIQ for different methods on the RealSR, RP60 and DIV2K datasets. We see that the curve of NSARM lies overall above the other methods, and it drops slowly at the end portion of the curve. In comparison, other methods drop very sharply, which implies that they will generate images of poor visual quality. The table in each sub-figure of Fig.~\ref{fig:robust} lists the numbers of Deficient, Poor and Collapse cases (refer to Eq. \ref{eq:robustness}) for each method. It can be observed that our model yields the fewest Deficient and Poor cases across all datasets, with no Collapse examples. SeeSR and SUPIR exhibit the most frequent Collapse failures because they generate HR images from Gaussian noise, which introduces more randomness. VARSR shows improved robustness through full finetuning (similar to our approach), yet remains limited by its non-LR starting point and discrete tokenization bottleneck, underperforming NSARM in both robustness and overall quality. This comprehensive validation confirms that NSARM has much higher robustness than other methods. More detailed distribution charts for each metric and dataset are in the \textbf{supplementary materials}.

\begin{table}[tbp!]
\centering
\adjustbox{width=\textwidth/2,center}{
\begin{tabular}{l|cc}
\toprule
Method & Inference Time (s) & Total Param (M) \\
\midrule
VARSR~\cite{qu2025visual} & 0.3 (512) & 1,102 \\
PURE~\cite{wei2025perceive} & 380 (512) & 7,080 \\
\midrule
OSEDiff~\cite{wu2024one} & 0.4 (1024) & 1,300 \\
PiSA-SR~\cite{sun2025pixel} & 0.4 (1024) & 1,775 \\
SeeSR~\cite{wu2024seesr} & 13 (1024) & 2,524 \\
SUPIR~\cite{yu2024scaling} & 11.7 (1024) & 4,801 \\
NSARM (ours) & 1.2 (1024) & 8,760 \\
\bottomrule
\end{tabular}
}
\caption{Complexity comparison of different methods. PURE and VARSR can only generate $512 \times 512$ images, while the others are evaluated on $1024 \times 1024$ resolution. The inference time is measured on an A800 80G GPU.}
\label{tab:complexity_comparison_vertical}
\vspace{-0.3cm}
\end{table}

\begin{figure*}[t]
\begin{center}
\includegraphics[width=1\linewidth]{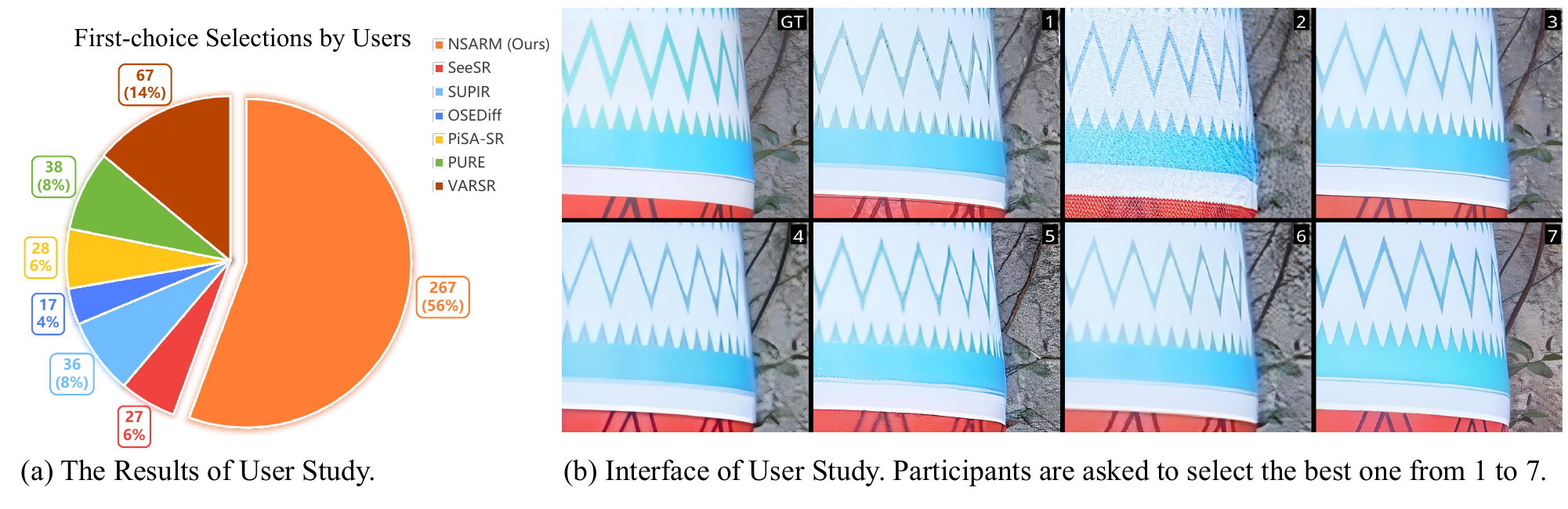}
\end{center}
\vspace{-0.5cm}
\caption{The results of user study. The participants are asked to select the best one from the 7 methods (the order of 1-7 was randomly changed in different test cases).}
\label{fig:user}
\vspace{-0.3cm}
\end{figure*}

\begin{figure*}[t]
\begin{center}
\includegraphics[width=1\linewidth]{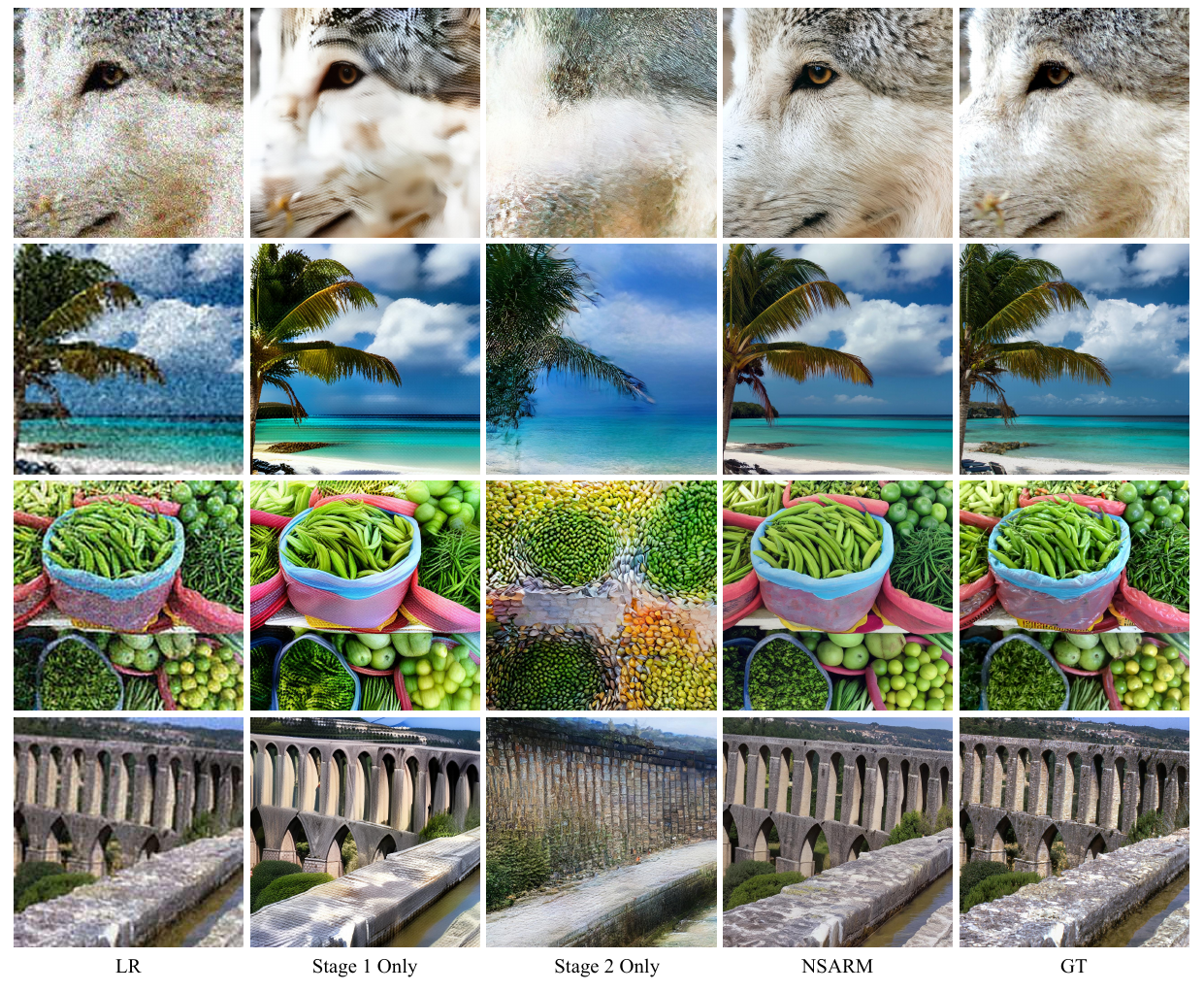}
\end{center}
\vspace{-0.3cm}
\caption{The comparison of NSARM with different training methods. Stage 1/2 Only: training the NSARM using only the first or second stage;  NSARM: training the model with the proposed two-stage optimization.}
\label{fig:training}
\vspace{-0.3cm}
\end{figure*}

\begin{figure}[t]
\begin{center}
\includegraphics[width=1\linewidth]{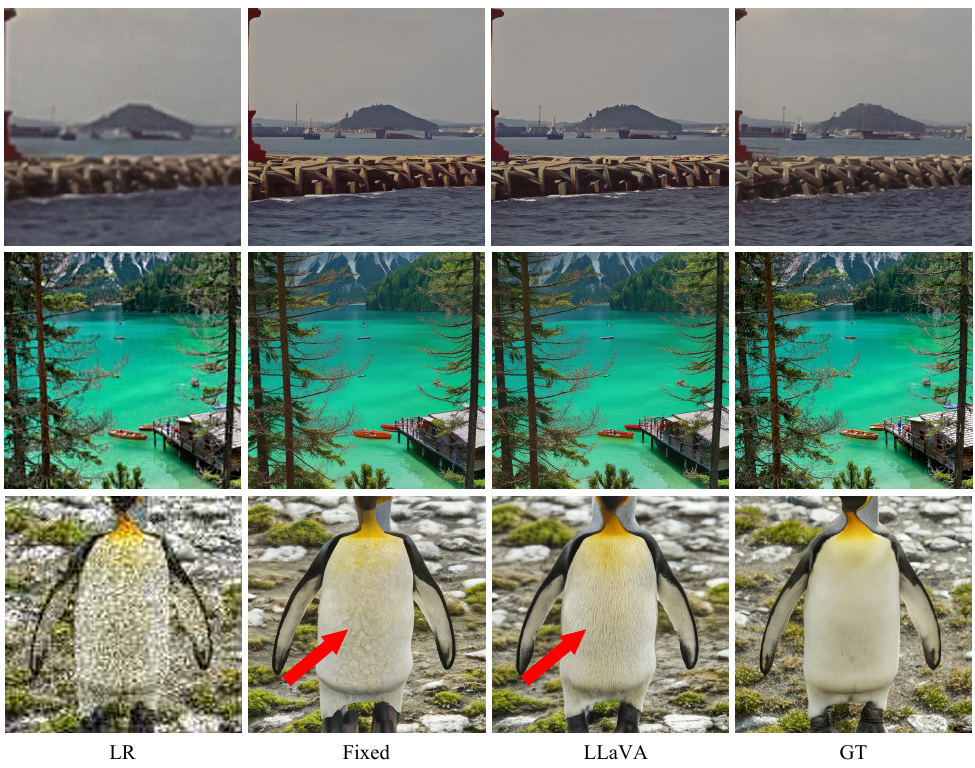}
\end{center}
\vspace{-0.3cm}
\caption{The comparison of NSARM with different text prompts. Fixed: using ``A high-quality image with harmonious colors and rich details."; LLaVA: using descriptions of image content extracted from the LR image by LLaVA.}
\label{fig:prompt}
\vspace{-0.3cm}
\end{figure}

\vspace{1mm}
\noindent\textbf{Complexity.}
Tab.~\ref{tab:complexity_comparison_vertical} compares the computational complexity of competing Real-ISR methods, including the number of parameters and inference time measured on a single A800 GPU. Note that the inference time of VARSR and PURE is measured at $512\times×512$ resolution, while others are measured at $1024\times1024$ resolution. NSARM demonstrates substantial inference speed advantages over iterative diffusion models, approximately 10$\times$ faster than SUPIR and SeeSR, while maintaining competitive speed with single-step methods, trailing OSEDiff and PiSASR by less than one second. Notably, our approach achieves over 100x acceleration compared to the pure AR method PURE, while matching the efficiency of VARSR (reported 0.3s is for $512\times×512$ generation). In terms of parameter scale, NSARM is not advantageous because of its foundation model backbone. Despite the large number of parameters, NSARM predicts next-scale tokens in bitwise space combined with infinite-vocabulary classifier technology, which enables efficient training and inference.


\vspace{1mm}
\noindent\textbf{Qualitative Comparisons.}
Visual comparisons with different Real-ISR methods are presented in Fig.~\ref{fig:visual}. In the first case, our approach reconstructs the most natural facial features and hair textures with exceptional clarity, while the competing methods exhibit significant shortcomings: SUPIR introduces over-sharp artifacts, and OSEDiff and PiSA-SR produce noticeable ghosting effects. The second case demonstrates our method's superior detail recovery capability: NSARM achieves photorealistic restoration and can handle the input with very low quality. The third example demonstrates the importance of model robustness: with challenging degradations, many methods (particularly those based on SD) will collapse, while NSARM maintains robust performance for those challenging images. Actually, due to the limitations of current IQA metrics for generative ISR, numerically similar scores may mask the substantial perceptual differences. More visual comparisons are provided in the \textbf{supplementary materials} to demonstrate the robustness of NSARM.

\subsection{User Study}

To comprehensively evaluate the perceptual quality, we conducted a user study with 12 participants. From each of the benchmark datasets (DIV2K~\cite{div2k}, DRealSR~\cite{drealsr}, RealSR~\cite{realsr}, and RP60~\cite{yu2024scaling}), we randomly selected 10 representative images (40 total) and generated the outputs of  the compared methods. Participants performed blind evaluations by selecting the best result from anonymized options (Fig.~\ref{fig:user}), with GT provided as the reference when available. Across the 480 total votes (12 participants × 40 images), our method received 267 first-choice selections (55.6\%), demonstrating clear user preference (Fig.~\ref{fig:user}). VARSR ranked second (14\% votes), benefiting from its VAR framework. OSEDiff and PiSA-SR (single-step diffusion) performed weaker (8\% and 6\% respectively), despite competitive numerical metrics. This reveals that their speed advantage comes at the cost of perceptual quality. 

\subsection{Ablation Study}
\label{sec:abla}

\paragraph{Necessity of Two-Stage Optimization.}

We present qualitative examples in Fig.~\ref{fig:training} to show the necessity of our two-stage optimization. As discussed in  Sec.~\ref{sec:The Pipeline of NSARM}, training the model using only Stage 1 or Stage 2 introduces artifacts or exhibits slow convergence. When solely training the Transformation Network without fine-tuning the full model (Stage 1 only), the output images retain the structure and content of the LR inputs but suffer from severe blurring and artifacts. This is because the Stage 1 training cannot achieve an optimal mapping that matches the subsequent prediction, leading to accumulated errors. Conversely, skipping Stage 1 and proceeding directly to Stage 2 can only roughly approximate the color distribution of the LR image. Without proper initialization from Stage 1, using the same number of iterations as in fine-tuning is insufficient for the model (from scratch) to converge. While the trend suggests that Stage 2 alone could eventually achieve good results, it requires prohibitively long training time. Our two-stage training combines the initialization in the first stage with the fine-tuning in the second stage, ensuring both the quality of reconstruction and the efficiency of convergence.

\begin{table*}[tb]
    \centering

    \adjustbox{width=\textwidth,center}{
\begin{tabular}{l|l|cccccccccccc}
    \toprule
    \multirow{2}{*}{Datasets} & \multirow{2}{*}{Methods} & \multirow{2}{*}{PSNR↑} & \multirow{2}{*}{SSIM↑} & \multirow{2}{*}{LPIPS↓} & \multirow{2}{*}{NIQE↓} & \multicolumn{2}{c}{MANIQA} & \multicolumn{2}{c}{MUSIQ} & \multicolumn{2}{c}{CLIPIQA}  & \multicolumn{2}{c}{TOPIQ}\\
     & & & & & & Avg.↑ & Var.↓ & Avg.↑ & Var.↓ & Avg.↑ & Var.↓ & Avg.↑ & Var.↓ \\
    \midrule
         \multicolumn{14}{c}{{RealESRGAN Degradations (Evaluating on generating 1024 $\times$ 1024 images)}} \\
    \midrule
         \multirow{2}{*}{DIV2K} 
       & NSARM (Fixed) & \textcolor{red}{20.17} & \textcolor{red}{0.4958} & 0.3761 & 4.07 & 0.6002 & \textcolor{red}{0.0036} & 73.26 & 11.2207 & 0.7129 & 0.0092 & 0.6648 & 0.0032   \\
      & NSARM (LLaVA) & \textcolor{red}{20.17} & 0.4950 & \textcolor{red}{0.3644} & \textcolor{red}{3.93} & \textcolor{red}{0.6094} & \textcolor{red}{0.0036} & \textcolor{red}{73.50} & \textcolor{red}{10.1565} & \textcolor{red}{0.7206} & \textcolor{red}{0.0082} & \textcolor{red}{0.6689} & \textcolor{red}{0.0025} \\
    \midrule
         \multicolumn{14}{c}{{Real Datasets (Evaluating on generating 512 $\times$ 512 images)}} \\
    \midrule
         \multirow{2}{*}{DRealSR} 
 & NSARM (Fixed) & 26.52 & 0.6889 & 0.4113 & 7.78 & 0.5756 & 0.0076 & 67.43 & 93.0262 & 0.7155 & 0.0123 & 0.6926 & 0.0064   \\
 & NSARM (LLaVA) & \textcolor{red}{26.66} & \textcolor{red}{0.6914} & \textcolor{red}{0.4038} & \textcolor{red}{7.56} & \textcolor{red}{0.5903} & \textcolor{red}{0.0069} & \textcolor{red}{67.67} & \textcolor{red}{89.8052} & \textcolor{red}{0.7259} & \textcolor{red}{0.0103} & \textcolor{red}{0.6991} & \textcolor{red}{0.0054} \\
    \midrule
         \multirow{2}{*}{RealSR} 
 & NSARM (Fixed) & \textcolor{red}{23.59} & \textcolor{red}{0.6436} & 0.3751 & 6.89 & 0.6356 & 0.0043 & 71.05 & 28.7962 & 0.7315 & \textcolor{red}{0.0061} & \textcolor{red}{0.7326} & 0.0017  \\
 & NSARM (LLaVA) & 23.52 & 0.6399 & \textcolor{red}{0.3706} & \textcolor{red}{6.81} & \textcolor{red}{0.6402} & \textcolor{red}{0.0037} & \textcolor{red}{71.09} & \textcolor{red}{24.5262} & \textcolor{red}{0.7370} & \textcolor{red}{0.0061} & {0.7318} & \textcolor{red}{0.0016} \\
    \midrule
         \multirow{2}{*}{RP60} 
      & NSARM (Fixed) & - & - & - & 5.74 & 0.6815 & 0.0029 & 73.85 & 12.1862 & \textcolor{red}{0.8596} & \textcolor{red}{0.0036} & 0.7583 & 0.0023  \\
      & NSARM (LLaVA) & - &-& - & \textcolor{red}{5.69} & \textcolor{red}{0.6862} & \textcolor{red}{0.0027} & \textcolor{red}{73.95} & \textcolor{red}{11.5072} & {0.8564} & {0.0040} & \textcolor{red}{0.7615} & \textcolor{red}{0.0021} \\
    \bottomrule
    \end{tabular} 

    }
\caption{Performance comparison of different NSARM variants on commonly used datasets. Fixed: using prompt ``A high-quality image with harmonious colors and rich details" during inference. LLaVA: using prompt extracted from the LR image by LLaVA~\cite{LLaVA}. \textcolor{red}{Red} color represents better performance. ↓ and  ↑ represent that the smaller or bigger is better.}
\label{tab:prompt}
\vspace{-0.3cm}
\end{table*}

\vspace{1mm}
\noindent\textbf{Impact of Text Prompt.}
As illustrated in Fig.~\ref{fig:main}, our model accepts text inputs, which can be content descriptions extracted from the LR image by a multimodal model (\eg, LLaVA~\cite{LLaVA}) or a fixed default prompt. For comparative analysis, we present two variants of NSARM outputs in Fig.~\ref{fig:prompt} and Tab.~\ref{tab:prompt}: (1) ``Fixed", which consistently employs a simple prompt ``A high-quality image with harmonious colors and rich details," and (2) ``LLaVA", where dynamic descriptions of image content are generated by processing the LR image with the LLaVA model. 

Quantitative results in Tab.~\ref{tab:prompt} demonstrate that, while LLaVA-generated prompts generally yield better performance, the improvement over fixed prompts is marginal in most cases. The visual comparisons in Fig.~\ref{fig:prompt} further corroborate this observation. As shown in the first two rows, both prompt variants achieve comparable reconstruction quality, successfully restoring even distant objects like ships. However, we note that precise prompts become more beneficial when the input image contains ambiguous local regions (\eg, the partial penguin view in the third row), where content-aware descriptions help guide more accurate generation. This relatively weak dependence on prompt specificity may be attributed to our scale replacement strategy. Note that the first scale of Infinity is derived from text input, and the prompts mainly influence the early-scale generations, which have been replaced by our inputs. Our results suggest that NSARM's performance is robust across most scenarios, regardless of prompt precision; even simple fixed prompts can produce visually satisfactory results.




\section{Conclusion}
\label{sec:Conclusion}
We presented NSARM, a robust and efficient framework for Real-ISR tasks, which effectively integrated the generative prior of the pre-trained Infinity model with a novel architectural design. Using bitwise next-scale prediction, NSARM achieved efficient, high-quality reconstruction through progressive refinement from LR inputs to HR outputs. Our two-stage training strategy, replacing preliminary scale by LR via a transformation network followed by full-parameter finetuning, ensured exceptional robustness while preserving strong generative capabilities. Extensive experiments demonstrated that NSARM achieves state-of-the-art visual quality with near-zero collapse cases across diverse scenarios, outperforming existing methods in both perceptual metrics and model robustness. Our work not only established a new framework for Real-ISR, but also validated that pure autoregressive models can be used to build promising paradigms for low-level vision tasks. 

\vspace{+2mm}
\noindent\textbf{Limitations.}
As an early exploration of autoregressive modeling for Real-ISR, NSARM also exhibits some limitations. (1) The bitwise generation paradigm reduces pixel-level fidelity compared to continuous-space alternatives, though without compromise of perceptual quality. (2) Current implementation adheres to the standard $1024\times1024$ output size of our base model, Infinity. However, this problem can be solved by patch-based inference or using some resizing operations in practice.

{
    \small
    \bibliographystyle{ieeenat_fullname}
    \bibliography{main}

\begin{thebibliography}{45}
\providecommand{\natexlab}[1]{#1}
\providecommand{\url}[1]{\texttt{#1}}
\expandafter\ifx\csname urlstyle\endcsname\relax
  \providecommand{\doi}[1]{doi: #1}\else
  \providecommand{\doi}{doi: \begingroup \urlstyle{rm}\Url}\fi

\bibitem[Agustsson and Timofte(2017)]{div2k}
Eirikur Agustsson and Radu Timofte.
\newblock Ntire 2017 challenge on single image super-resolution: Dataset and study.
\newblock In \emph{Proceedings of the IEEE conference on computer vision and pattern recognition workshops}, pages 126--135, 2017.

\bibitem[Cai et~al.(2019)Cai, Zeng, Yong, Cao, and Zhang]{realsr}
Jianrui Cai, Hui Zeng, Hongwei Yong, Zisheng Cao, and Lei Zhang.
\newblock Toward real-world single image super-resolution: A new benchmark and a new model.
\newblock In \emph{Proceedings of the IEEE/CVF International Conference on Computer Vision}, pages 3086--3095, 2019.

\bibitem[Chen et~al.(2024)Chen, Mo, Hou, Wu, Liao, Sun, Yan, and Lin]{chen2024topiq}
Chaofeng Chen, Jiadi Mo, Jingwen Hou, Haoning Wu, Liang Liao, Wenxiu Sun, Qiong Yan, and Weisi Lin.
\newblock Topiq: A top-down approach from semantics to distortions for image quality assessment.
\newblock \emph{IEEE Transactions on Image Processing}, 33:\penalty0 2404--2418, 2024.

\bibitem[Chen et~al.(2023)Chen, Wang, Zhang, Kong, Qiao, Zhou, and Dong]{chen2023hat}
Xiangyu Chen, Xintao Wang, Wenlong Zhang, Xiangtao Kong, Yu Qiao, Jiantao Zhou, and Chao Dong.
\newblock Hat: Hybrid attention transformer for image restoration.
\newblock \emph{arXiv preprint arXiv:2309.05239}, 2023.

\bibitem[Dai et~al.(2019)Dai, Cai, Zhang, Xia, and Zhang]{dai2019second}
Tao Dai, Jianrui Cai, Yongbing Zhang, Shu-Tao Xia, and Lei Zhang.
\newblock Second-order attention network for single image super-resolution.
\newblock In \emph{Proceedings of the IEEE/CVF conference on computer vision and pattern recognition}, pages 11065--11074, 2019.

\bibitem[Dong et~al.(2014)Dong, Loy, He, and Tang]{dong2014learning}
Chao Dong, Chen~Change Loy, Kaiming He, and Xiaoou Tang.
\newblock Learning a deep convolutional network for image super-resolution.
\newblock In \emph{Computer Vision--ECCV 2014: 13th European Conference, Zurich, Switzerland, September 6-12, 2014, Proceedings, Part IV 13}, pages 184--199. Springer, 2014.

\bibitem[Dong et~al.(2015)Dong, Loy, He, and Tang]{dong2015image}
Chao Dong, Chen~Change Loy, Kaiming He, and Xiaoou Tang.
\newblock Image super-resolution using deep convolutional networks.
\newblock \emph{IEEE transactions on pattern analysis and machine intelligence}, 38\penalty0 (2):\penalty0 295--307, 2015.

\bibitem[Esser et~al.(2021{\natexlab{a}})Esser, Rombach, and Ommer]{esser2021taming}
Patrick Esser, Robin Rombach, and Bjorn Ommer.
\newblock Taming transformers for high-resolution image synthesis.
\newblock In \emph{Proceedings of the IEEE/CVF conference on computer vision and pattern recognition}, pages 12873--12883, 2021{\natexlab{a}}.

\bibitem[Esser et~al.(2021{\natexlab{b}})Esser, Rombach, and Ommer]{vqgan}
Patrick Esser, Robin Rombach, and Bjorn Ommer.
\newblock Taming transformers for high-resolution image synthesis.
\newblock In \emph{Proceedings of the IEEE/CVF conference on computer vision and pattern recognition}, pages 12873--12883, 2021{\natexlab{b}}.

\bibitem[Esser et~al.(2024)Esser, Kulal, Blattmann, Entezari, M{\"u}ller, Saini, Levi, Lorenz, Sauer, Boesel, et~al.]{stable-diffusion3}
Patrick Esser, Sumith Kulal, Andreas Blattmann, Rahim Entezari, Jonas M{\"u}ller, Harry Saini, Yam Levi, Dominik Lorenz, Axel Sauer, Frederic Boesel, et~al.
\newblock Scaling rectified flow transformers for high-resolution image synthesis.
\newblock In \emph{Forty-first International Conference on Machine Learning}, 2024.

\bibitem[Goodfellow et~al.(2020)Goodfellow, Pouget-Abadie, Mirza, Xu, Warde-Farley, Ozair, Courville, and Bengio]{goodfellow2020generative}
Ian Goodfellow, Jean Pouget-Abadie, Mehdi Mirza, Bing Xu, David Warde-Farley, Sherjil Ozair, Aaron Courville, and Yoshua Bengio.
\newblock Generative adversarial networks.
\newblock \emph{Communications of the ACM}, 63\penalty0 (11):\penalty0 139--144, 2020.

\bibitem[Han et~al.(2024)Han, Liu, Jiang, Yan, Zhang, Yuan, Peng, and Liu]{han2024infinity}
Jian Han, Jinlai Liu, Yi Jiang, Bin Yan, Yuqi Zhang, Zehuan Yuan, Bingyue Peng, and Xiaobing Liu.
\newblock Infinity: Scaling bitwise autoregressive modeling for high-resolution image synthesis.
\newblock \emph{arXiv preprint arXiv:2412.04431}, 2024.

\bibitem[Ke et~al.(2021)Ke, Wang, Wang, Milanfar, and Yang]{musiq}
Junjie Ke, Qifei Wang, Yilin Wang, Peyman Milanfar, and Feng Yang.
\newblock Musiq: Multi-scale image quality transformer.
\newblock In \emph{Proceedings of the IEEE/CVF International Conference on Computer Vision}, pages 5148--5157, 2021.

\bibitem[Ledig et~al.(2017)Ledig, Theis, Husz{\'a}r, Caballero, Cunningham, Acosta, Aitken, Tejani, Totz, Wang, et~al.]{SRResNet}
Christian Ledig, Lucas Theis, Ferenc Husz{\'a}r, Jose Caballero, Andrew Cunningham, Alejandro Acosta, Andrew Aitken, Alykhan Tejani, Johannes Totz, Zehan Wang, et~al.
\newblock Photo-realistic single image super-resolution using a generative adversarial network.
\newblock In \emph{Proceedings of the IEEE conference on computer vision and pattern recognition}, pages 4681--4690, 2017.

\bibitem[Lin et~al.(2023)Lin, He, Chen, Lyu, Fei, Dai, Ouyang, Qiao, and Dong]{lin2023diffbir}
Xinqi Lin, Jingwen He, Ziyan Chen, Zhaoyang Lyu, Ben Fei, Bo Dai, Wanli Ouyang, Yu Qiao, and Chao Dong.
\newblock Diffbir: Towards blind image restoration with generative diffusion prior.
\newblock \emph{arXiv preprint arXiv:2308.15070}, 2023.

\bibitem[Liu et~al.(2024{\natexlab{a}})Liu, Zhao, Zhuo, Lin, Xin, Li, Qin, Qiao, Li, and Gao]{liu2024lumina}
Dongyang Liu, Shitian Zhao, Le Zhuo, Weifeng Lin, Yi Xin, Xinyue Li, Qi Qin, Yu Qiao, Hongsheng Li, and Peng Gao.
\newblock Lumina-mgpt: Illuminate flexible photorealistic text-to-image generation with multimodal generative pretraining.
\newblock \emph{arXiv preprint arXiv:2408.02657}, 2024{\natexlab{a}}.

\bibitem[Liu et~al.(2024{\natexlab{b}})Liu, Li, Li, and Lee]{LLaVA}
Haotian Liu, Chunyuan Li, Yuheng Li, and Yong~Jae Lee.
\newblock Improved baselines with visual instruction tuning.
\newblock In \emph{Proceedings of the IEEE/CVF conference on computer vision and pattern recognition}, pages 26296--26306, 2024{\natexlab{b}}.

\bibitem[Loshchilov and Hutter(2017)]{adamw}
Ilya Loshchilov and Frank Hutter.
\newblock Decoupled weight decay regularization.
\newblock \emph{arXiv preprint arXiv:1711.05101}, 2017.

\bibitem[Podell et~al.(2023)Podell, English, Lacey, Blattmann, Dockhorn, M{\"u}ller, Penna, and Rombach]{sdxl}
Dustin Podell, Zion English, Kyle Lacey, Andreas Blattmann, Tim Dockhorn, Jonas M{\"u}ller, Joe Penna, and Robin Rombach.
\newblock Sdxl: Improving latent diffusion models for high-resolution image synthesis.
\newblock \emph{arXiv preprint arXiv:2307.01952}, 2023.

\bibitem[Qu et~al.(2025)Qu, Yuan, Hao, Zhao, Xie, Sun, and Zhou]{qu2025visual}
Yunpeng Qu, Kun Yuan, Jinhua Hao, Kai Zhao, Qizhi Xie, Ming Sun, and Chao Zhou.
\newblock Visual autoregressive modeling for image super-resolution.
\newblock \emph{arXiv preprint arXiv:2501.18993}, 2025.

\bibitem[{Radford} et~al.(2021){Radford}, {Kim}, {Hallacy}, {Ramesh}, {Goh}, {Agarwal}, {Sastry}, {Askell}, {Mishkin}, {Clark}, {Krueger}, and {Sutskever}]{CLIP}
Alec {Radford}, Jong~Wook {Kim}, Chris {Hallacy}, Aditya {Ramesh}, Gabriel {Goh}, Sandhini {Agarwal}, Girish {Sastry}, Amanda {Askell}, Pamela {Mishkin}, Jack {Clark}, Gretchen {Krueger}, and Ilya {Sutskever}.
\newblock {Learning Transferable Visual Models From Natural Language Supervision}.
\newblock \emph{arXiv e-prints}, art. arXiv:2103.00020, 2021.

\bibitem[Ramesh et~al.(2021)Ramesh, Pavlov, Goh, Gray, Voss, Radford, Chen, and Sutskever]{dalle1}
Aditya Ramesh, Mikhail Pavlov, Gabriel Goh, Scott Gray, Chelsea Voss, Alec Radford, Mark Chen, and Ilya Sutskever.
\newblock Zero-shot text-to-image generation.
\newblock In \emph{International Conference on Machine Learning}, pages 8821--8831. PMLR, 2021.

\bibitem[Rombach et~al.(2022)Rombach, Blattmann, Lorenz, Esser, and Ommer]{sd2.1}
Robin Rombach, Andreas Blattmann, Dominik Lorenz, Patrick Esser, and Bj{\"o}rn Ommer.
\newblock High-resolution image synthesis with latent diffusion models.
\newblock In \emph{Proceedings of the IEEE/CVF conference on computer vision and pattern recognition}, pages 10684--10695, 2022.

\bibitem[Sun et~al.(2025)Sun, Wu, Ma, Liu, Yi, and Zhang]{sun2025pixel}
Lingchen Sun, Rongyuan Wu, Zhiyuan Ma, Shuaizheng Liu, Qiaosi Yi, and Lei Zhang.
\newblock Pixel-level and semantic-level adjustable super-resolution: A dual-lora approach.
\newblock In \emph{Proceedings of the Computer Vision and Pattern Recognition Conference}, pages 2333--2343, 2025.

\bibitem[Tian et~al.(2024)Tian, Jiang, Yuan, Peng, and Wang]{VAR}
Keyu Tian, Yi Jiang, Zehuan Yuan, Bingyue Peng, and Liwei Wang.
\newblock Visual autoregressive modeling: Scalable image generation via next-scale prediction.
\newblock \emph{arXiv preprint arXiv:2404.02905}, 2024.

\bibitem[Unsplash-website()]{unsplash}
Unsplash-website.
\newblock Unsplash-website.
\newblock \url{https://unsplash.com/data}.

\bibitem[Wang et~al.(2023)Wang, Chan, and Loy]{clipiqa}
Jianyi Wang, Kelvin~CK Chan, and Chen~Change Loy.
\newblock Exploring clip for assessing the look and feel of images.
\newblock In \emph{Proceedings of the AAAI Conference on Artificial Intelligence}, pages 2555--2563, 2023.

\bibitem[Wang et~al.(2024)Wang, Yue, Zhou, Chan, and Loy]{wang2024exploiting}
Jianyi Wang, Zongsheng Yue, Shangchen Zhou, Kelvin~CK Chan, and Chen~Change Loy.
\newblock Exploiting diffusion prior for real-world image super-resolution.
\newblock \emph{International Journal of Computer Vision}, 132\penalty0 (12):\penalty0 5929--5949, 2024.

\bibitem[Wang et~al.(2018)Wang, Yu, Wu, Gu, Liu, Dong, Qiao, and Change~Loy]{wang2018esrgan}
Xintao Wang, Ke Yu, Shixiang Wu, Jinjin Gu, Yihao Liu, Chao Dong, Yu Qiao, and Chen Change~Loy.
\newblock Esrgan: Enhanced super-resolution generative adversarial networks.
\newblock In \emph{Proceedings of the European conference on computer vision (ECCV) workshops}, pages 0--0, 2018.

\bibitem[Wang et~al.(2021)Wang, Xie, Dong, and Shan]{wang2021real}
Xintao Wang, Liangbin Xie, Chao Dong, and Ying Shan.
\newblock Real-esrgan: Training real-world blind super-resolution with pure synthetic data.
\newblock In \emph{Proceedings of the IEEE/CVF international conference on computer vision}, pages 1905--1914, 2021.

\bibitem[Wei et~al.(2025)Wei, Liu, Yuan, and Zhang]{wei2025perceive}
Hongyang Wei, Shuaizheng Liu, Chun Yuan, and Lei Zhang.
\newblock Perceive, understand and restore: Real-world image super-resolution with autoregressive multimodal generative models.
\newblock \emph{arXiv preprint arXiv:2503.11073}, 2025.

\bibitem[Wei et~al.(2020)Wei, Xie, Lu, Zhan, Ye, Zuo, and Lin]{drealsr}
Pengxu Wei, Ziwei Xie, Hannan Lu, Zongyuan Zhan, Qixiang Ye, Wangmeng Zuo, and Liang Lin.
\newblock Component divide-and-conquer for real-world image super-resolution.
\newblock In \emph{Computer Vision--ECCV 2020: 16th European Conference, Glasgow, UK, August 23--28, 2020, Proceedings, Part VIII 16}, pages 101--117. Springer, 2020.

\bibitem[Wu et~al.(2024{\natexlab{a}})Wu, Sun, Ma, and Zhang]{wu2024one}
Rongyuan Wu, Lingchen Sun, Zhiyuan Ma, and Lei Zhang.
\newblock One-step effective diffusion network for real-world image super-resolution.
\newblock \emph{Advances in Neural Information Processing Systems}, 37:\penalty0 92529--92553, 2024{\natexlab{a}}.

\bibitem[Wu et~al.(2024{\natexlab{b}})Wu, Yang, Sun, Zhang, Li, and Zhang]{wu2024seesr}
Rongyuan Wu, Tao Yang, Lingchen Sun, Zhengqiang Zhang, Shuai Li, and Lei Zhang.
\newblock Seesr: Towards semantics-aware real-world image super-resolution.
\newblock In \emph{Proceedings of the IEEE/CVF conference on computer vision and pattern recognition}, pages 25456--25467, 2024{\natexlab{b}}.

\bibitem[Yang et~al.(2022)Yang, Wu, Shi, Lao, Gong, Cao, Wang, and Yang]{maniqa}
Sidi Yang, Tianhe Wu, Shuwei Shi, Shanshan Lao, Yuan Gong, Mingdeng Cao, Jiahao Wang, and Yujiu Yang.
\newblock Maniqa: Multi-dimension attention network for no-reference image quality assessment.
\newblock In \emph{Proceedings of the IEEE/CVF Conference on Computer Vision and Pattern Recognition}, pages 1191--1200, 2022.

\bibitem[Yang et~al.(2023)Yang, Ren, Xie, and Zhang]{yang2023pixel}
Tao Yang, Peiran Ren, Xuansong Xie, and Lei Zhang.
\newblock Pixel-aware stable diffusion for realistic image super-resolution and personalized stylization.
\newblock \emph{arXiv preprint arXiv:2308.14469}, 2023.

\bibitem[Yang et~al.(2024)Yang, Wu, Ren, Xie, and Zhang]{yang2024pixel}
Tao Yang, Rongyuan Wu, Peiran Ren, Xuansong Xie, and Lei Zhang.
\newblock Pixel-aware stable diffusion for realistic image super-resolution and personalized stylization.
\newblock In \emph{European Conference on Computer Vision}, pages 74--91. Springer, 2024.

\bibitem[Yu et~al.(2024)Yu, Gu, Li, Hu, Kong, Wang, He, Qiao, and Dong]{yu2024scaling}
Fanghua Yu, Jinjin Gu, Zheyuan Li, Jinfan Hu, Xiangtao Kong, Xintao Wang, Jingwen He, Yu Qiao, and Chao Dong.
\newblock Scaling up to excellence: Practicing model scaling for photo-realistic image restoration in the wild.
\newblock In \emph{Proceedings of the IEEE/CVF Conference on Computer Vision and Pattern Recognition}, pages 25669--25680, 2024.

\bibitem[Yu et~al.(2022)Yu, Xu, Koh, Luong, Baid, Wang, Vasudevan, Ku, Yang, Ayan, et~al.]{parti}
Jiahui Yu, Yuanzhong Xu, Jing~Yu Koh, Thang Luong, Gunjan Baid, Zirui Wang, Vijay Vasudevan, Alexander Ku, Yinfei Yang, Burcu~Karagol Ayan, et~al.
\newblock Scaling autoregressive models for content-rich text-to-image generation.
\newblock \emph{arXiv preprint arXiv:2206.10789}, 2\penalty0 (3):\penalty0 5, 2022.

\bibitem[Yue et~al.(2023)Yue, Wang, and Loy]{yue2023resshift}
Zongsheng Yue, Jianyi Wang, and Chen~Change Loy.
\newblock Resshift: Efficient diffusion model for image super-resolution by residual shifting.
\newblock \emph{arXiv preprint arXiv:2307.12348}, 2023.

\bibitem[Zhang et~al.(2021)Zhang, Liang, Van~Gool, and Timofte]{zhang2021designing}
Kai Zhang, Jingyun Liang, Luc Van~Gool, and Radu Timofte.
\newblock Designing a practical degradation model for deep blind image super-resolution.
\newblock In \emph{Proceedings of the IEEE/CVF International Conference on Computer Vision}, pages 4791--4800, 2021.

\bibitem[Zhang et~al.(2015)Zhang, Zhang, and Bovik]{niqe}
Lin Zhang, Lei Zhang, and Alan~C Bovik.
\newblock A feature-enriched completely blind image quality evaluator.
\newblock \emph{IEEE Transactions on Image Processing}, 24\penalty0 (8):\penalty0 2579--2591, 2015.

\bibitem[Zhang et~al.(2018)Zhang, Isola, Efros, Shechtman, and Wang]{lpips}
Richard Zhang, Phillip Isola, Alexei~A Efros, Eli Shechtman, and Oliver Wang.
\newblock The unreasonable effectiveness of deep features as a perceptual metric.
\newblock In \emph{Proceedings of the IEEE conference on computer vision and pattern recognition}, pages 586--595, 2018.

\bibitem[Zhang et~al.(2019)Zhang, Liu, Dong, and Qiao]{zhang2019ranksrgan}
Wenlong Zhang, Yihao Liu, Chao Dong, and Yu Qiao.
\newblock Ranksrgan: Generative adversarial networks with ranker for image super-resolution.
\newblock In \emph{Proceedings of the IEEE/CVF international conference on computer vision}, pages 3096--3105, 2019.

\bibitem[Zhao et~al.(2024)Zhao, Xiong, and Kr{\"a}henb{\"u}hl]{zhao2024image}
Yue Zhao, Yuanjun Xiong, and Philipp Kr{\"a}henb{\"u}hl.
\newblock Image and video tokenization with binary spherical quantization.
\newblock \emph{arXiv preprint arXiv:2406.07548}, 2024.

\end{thebibliography}
}

\clearpage
\renewcommand\thesection{\Alph{section}}
	\renewcommand\thesubsection{\thesection.\arabic{subsection}}
	\renewcommand\thefigure{\Alph{section}.\arabic{figure}}
	\renewcommand\thetable{\Alph{section}.\arabic{table}}

\twocolumn[%
	\noindent
	\begin{@twocolumnfalse}
		\renewcommand\thesection{\Alph{section}}
		\renewcommand\thesubsection{\thesection.\arabic{subsection}}
		\renewcommand\thefigure{\Alph{section}.\arabic{figure}}
		\renewcommand\thetable{\Alph{section}.\arabic{table}} 
		
		\begin{center}
			\LARGE\textbf{NSARM: Next-Scale Autoregressive Modeling for Robust \\ Real-World Image Super-Resolution 
            \\ - Supplementary Materials -}
		\end{center}
		\vspace{1em}
	\end{@twocolumnfalse}
]
	
	\setcounter{section}{0}
	\setcounter{figure}{0}
	\setcounter{table}{0}

In the supplementary materials, we first present the pseudo code of our proposed algorithm, and then provide additional experimental results, including more visual and quantitative comparisons, as well as more sorted metric score distribution curves.

\section{Pseudo Code of NSARM}

To facilitate the understanding of our algorithm, we provide the pseudo code of NSARM, which offers a more comprehensive view of the implementation of our method. First, we show the Visual Tokenizer Encoding (how to obtain residuals from GT) in Alg.~\ref{alg:algorithm1}. Latent features $\bm{F}$ are obtained from the encoding of GT, which could be decomposed into the residual queue $\bm{R}_{queue}$ and accumulated $\widetilde{\bm{F}}_{queue}$. In the original training of Infinity, $\widetilde{\bm{F}}_{queue}$ contains the inputs of each scale (to predict the residual of the next scale) and $\bm{R}_{queue}$ includes the corresponding residual labels. Our proposed transformation network $\mathcal{T}(\cdot)$ transforms LR to preliminary scale residuals (1 to 7 elements of $\bm{R}_{queue}$), which is optimized by our Stage 1 training.

\begin{algorithm}[h]
\caption{Visual Tokenizer Encoding}
\label{alg:algorithm1}
\textbf{Input}: Raw feature $\bm{F}$, \\scale schedule $\{(h^r_1,w^r_1),...,(h^r_K,w^r_K)\}$\\
\textbf{Output}: $\bm{R}_{queue}, \widetilde{\bm{F}}_{queue}$ 
\begin{algorithmic}[1] 
\STATE $\bm{R}_{queue}=[]$ (multi-scale bit labels)
\STATE $\widetilde{\bm{F}}_{queue}=[]$ (inputs for AR, accumulated $\bm{R}$)
\FOR {$k=1,2,\cdots,K \vphantom{\bm{F}^{flip}_{k-1}}$}
\STATE $\bm{R}_k=\mathcal{Q}(\operatorname{down}(\bm{F} - \bm{F}_{k-1}, (h_{k}, w_{k}))$
\STATE $\operatorname{Queue\_Push}$($\bm{R}_{queue}, \bm{R}_{k}$)
\STATE $\bm{F}_{k} = \sum_{i=1}^{k} \operatorname{up}(\bm{R}_i, (h, w))$
\STATE $\widetilde{\bm{F}}_k = \operatorname{down}(\bm{F}_{k}, (h_{k+1}, w_{k+1}))$
\STATE $\operatorname{Queue\_Push}$($\widetilde{\bm{F}}_{queue}, \widetilde{\bm{F}}_k$)
\ENDFOR
\end{algorithmic}
\end{algorithm}

Then, we obtain the transformed $\bm{R'}_{queue} = \{\mathcal{T}(LR)_1, \mathcal{T}(LR)_2, \dots, \mathcal{T}(LR)_7\}$, which are used to replace the original preliminary scale residuals. During Stage 2 training, NSARM employs modified GT residual labels to constrain the generation pathway from $\bm{R'}_{queue}$ to GT. This fine-tuning process effectively repositions $\bm{R'}_{queue}$ as the new starting point for the generation. As illustrated in Alg.~\ref{alg:algorithm2}, the replacement of preliminary scale residuals induces a cascaded modification throughout the entire sequence of components $\bm{R}_{queue\_m}$ and $\widetilde{\bm{F}}_{queue\_m}$. This transformation process begins with the transformed $\bm{R'}_{queue}$, with each subsequent step progressively refined by the GT feature term $F$. This produces the supervision signal ($\bm{R}_{queue\_m}$) from LR toward the desired HR, which is also the label of our Stage 2 training. Finally, the modified inputs and labels would be constrained by Eq.~\ref{eq:ce_loss} to execute the training of Stage 2.

\begin{algorithm}[t]
\caption{Cascaded GT Modification}
\label{alg:algorithm2}
\textbf{Input}: Raw feature $\bm{F}$, \\scale schedule $\{(h^r_1,w^r_1),...,(h^r_K,w^r_K)\}$,\\
transformed LR $\bm{R'}_{queue}$ \\
\textbf{Output}: $\bm{R}_{queue\_m}, \widetilde{\bm{F}}_{queue\_m}$ 
\begin{algorithmic}[1] 
\STATE $\bm{R}_{queue\_m}=[]$ (modified multi-scale bit labels)
\STATE $\widetilde{\bm{F}}_{queue\_m}=[]$ (modified inputs, accumulated $\bm{R}$)
\FOR {$k=1,2,\cdots,K \vphantom{\bm{F}^{flip}_{k-1}}$}
\IF {$k<=7$}
\STATE $\bm{R}_k=\bm{R'}_{queue}$
\ELSE
\STATE $\bm{R}_k=\mathcal{Q}(\operatorname{down}(\bm{F} - \bm{F}_{k-1}, (h_{k}, w_{k}))$
\ENDIF
\STATE $\operatorname{Queue\_Push}$($\bm{R}_{queue\_m}, \bm{R}_{k}$)
\STATE $\bm{F}_{k} = \sum_{i=1}^{k} \operatorname{up}(\bm{R}_i, (h, w))$
\STATE $\widetilde{\bm{F}}_k = \operatorname{down}(\bm{F}_{k}, (h_{k+1}, w_{k+1}))$
\STATE $\operatorname{Queue\_Push}$($\widetilde{\bm{F}}_{queue\_m}, \widetilde{\bm{F}}_k$)
\ENDFOR
\end{algorithmic}
\end{algorithm}

\section{More Experimental Results}

\begin{figure*}[t]
\begin{center}
\includegraphics[width=1\linewidth]{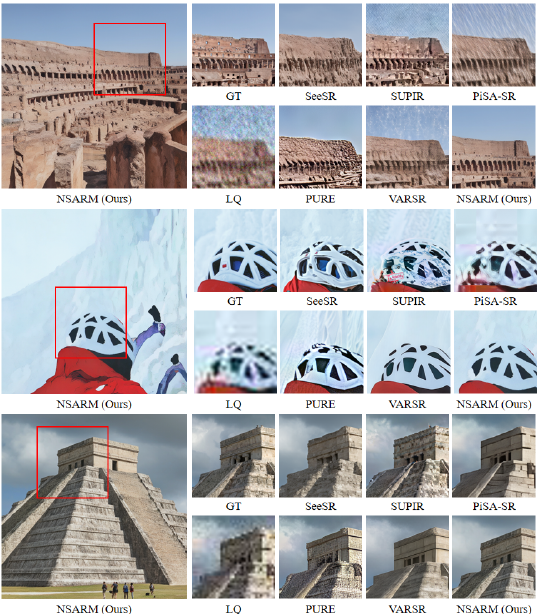}
\end{center}
\caption{Visual comparison of different methods on DIV2K (zoom in for a better view).}
\label{fig:visual1}
\end{figure*}

\begin{figure*}[t]
\begin{center}
\includegraphics[width=1\linewidth]{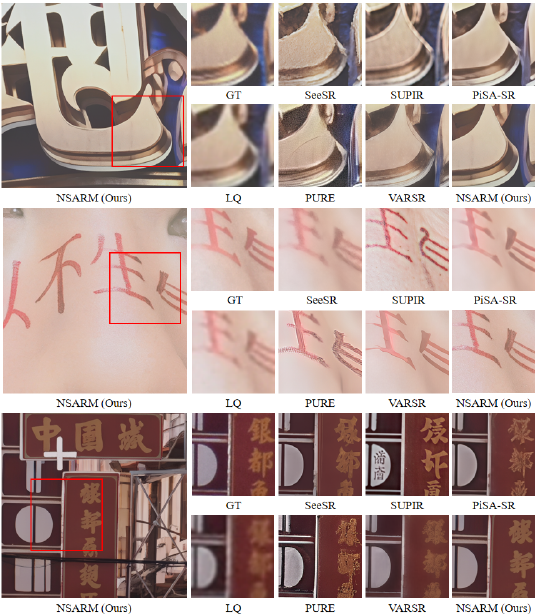}
\end{center}
\caption{Visual comparison of different methods on DRealSR (zoom in for a better view).}
\label{fig:visual2}
\end{figure*}

\begin{figure*}[t]
\begin{center}
\includegraphics[width=1\linewidth]{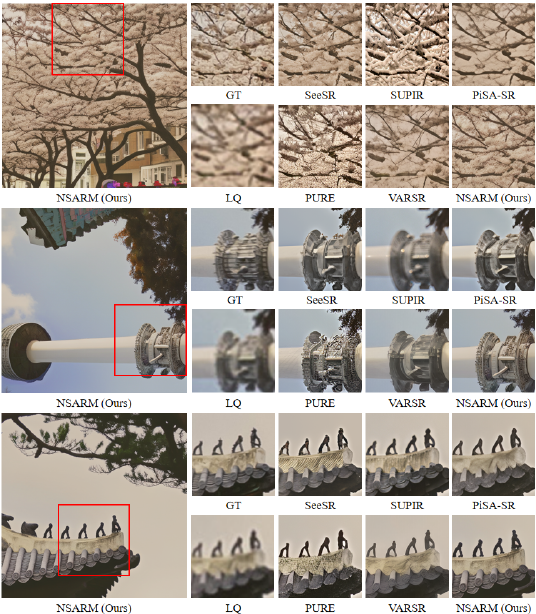}
\end{center}
\caption{Visual comparison of different methods on RealSR (zoom in for a better view).}
\label{fig:visual3}
\end{figure*}

\begin{figure*}[t]
\begin{center}
\includegraphics[width=1\linewidth]{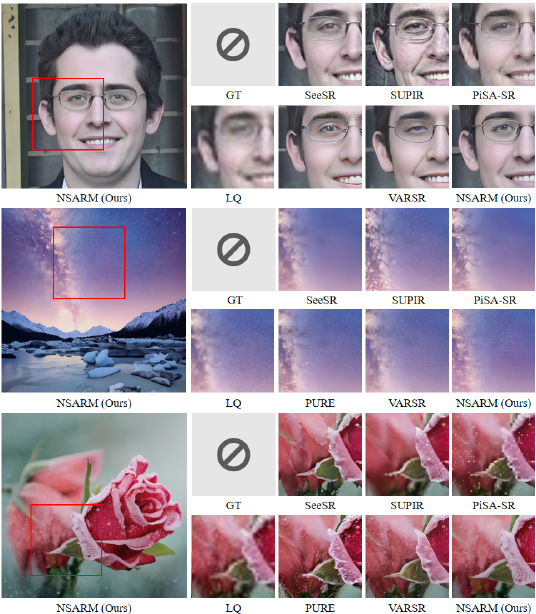}
\end{center}
\caption{Visual comparison of different methods on RP60 (zoom in for a better view).}
\label{fig:visual4}
\end{figure*}

\paragraph{More Visual Results.}
As mentioned in the section of Qualitative Comparisons in the main paper, current IQA metrics for Real-ISR have limitations, where numerically similar scores may obscure significant perceptual differences. Here, we provide additional visual comparisons to demonstrate NSARM's robustness. Figs.~\ref{fig:visual1} to~\ref{fig:visual4} present results on DIV2K~\cite{div2k}, DRealSR~\cite{drealsr}, RealSR~\cite{realsr}, and RP60~\cite{yu2024scaling} datasets, respectively. Note that, due to layout constraints and considering that PiSA-SR (as an enhanced version of OSEDiff) produces similar effects to OESDiff, we only present the result of PiSA-SR for comparison.

In Fig.~\ref{fig:visual1}, we see that our method consistently produces clearer visual results without introducing background artifacts. Unlike PURE or SUPIR, NSARM avoids the generation of over-sharpening or hallucinated details. Fig.~\ref{fig:visual2} demonstrates the superiority of our method in reconstructing clean and sharp text. Notably, even in challenging cases where all methods fail to recover the text (third example), our results remain the clearest and almost artifact-free while maintaining robustness against hallucinations. Fig.~\ref{fig:visual3} shows that the outputs of NSARM match closely the GT's visual quality: it accurately recovers cherry blossoms and branches in the first row (while SUPIR mistakenly interprets them as snow-covered trees). In the third row, NSARM preserves smooth white areas while correctly rendering tiled roof textures. The first row of Fig.~\ref{fig:visual4} highlights NSARM's advantages in facial and hair reconstruction. The second row shows that NSARM preserves star details without mistaking them for noise. Besides, NSARM successfully recovers ``frozen roses in snowy scenes" (third row) where many methods fail.

Our visual analysis reveals the distinct characteristics among the compared methods. SUPIR and PURE frequently produce over-sharpened results with hallucinated content, particularly in texture-rich regions (\eg, the first sample of Fig.~\ref{fig:visual1} and the last row of Fig.~\ref{fig:visual2}). PiSA-SR, as a single-stage diffusion model, exhibits limited generative capability, often yielding oversmoothed outputs or visible artifacts (\eg, the first sample of both Fig.~\ref{fig:visual3} and Fig.~\ref{fig:visual4}). Although VARSR achieves competitive numerical metrics (as reported in the main paper), its reconstructions tend to be excessively smooth. In contrast, our NSARM maintains a good balance, avoiding both the hallucination of SUPIR/PURE and the over-smoothing issues of PiSA-SR/VARSR, while consistently producing the most faithful reconstructions across most test cases.

\vspace{+1mm}
\noindent\textbf{More Quantitative Results.}
As mentioned in the Testing Protocol section of the main paper, while SeeSR, OSEDiff, and PiSA-SR were originally trained on $512 \times 512$ images, we compared them with them by first generating $1024 \times 1024$ outputs followed by downsampling to $512 \times 512$ for unified evaluation. In Tab.~\ref{tab:methods_comparison_512}, we additionally present results from these methods when directly generating $512 \times 512$ images. The results reveal two observations: (1) The ``first generating $1024 \times 1024$ then downsampling to $512 \times 512$" approach generally achieves better fidelity metrics, benefiting from the additional computational resources expended. (2) For perceptual metrics, performance varies, with some metrics improving while others degrade. Importantly, even when comparing our downsampled $512 \times 512$ results with natively trained models, NSARM maintains superior performance on most perceptual quality measures, demonstrating its robustness in qualified reconstruction.

\begin{figure*}[t]
\begin{center}
\includegraphics[width=1\linewidth]{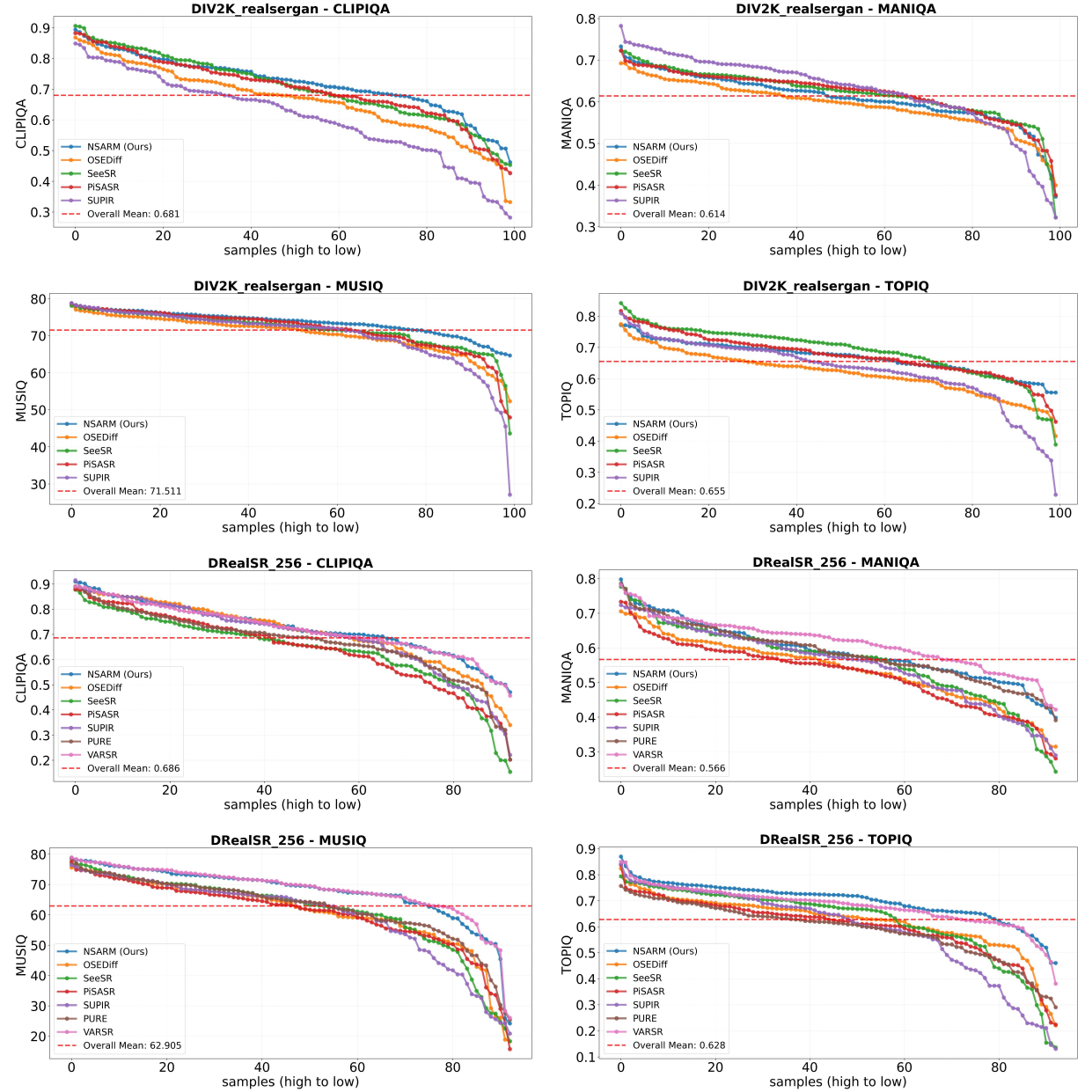}
\end{center}
\caption{Detailed ranked score distributions of Real-ISR methods across multiple evaluation metrics on DIV2K and DRealSR.}
\label{fig:dis1}
\end{figure*}

\begin{figure*}[t]
\begin{center}
\includegraphics[width=1\linewidth]{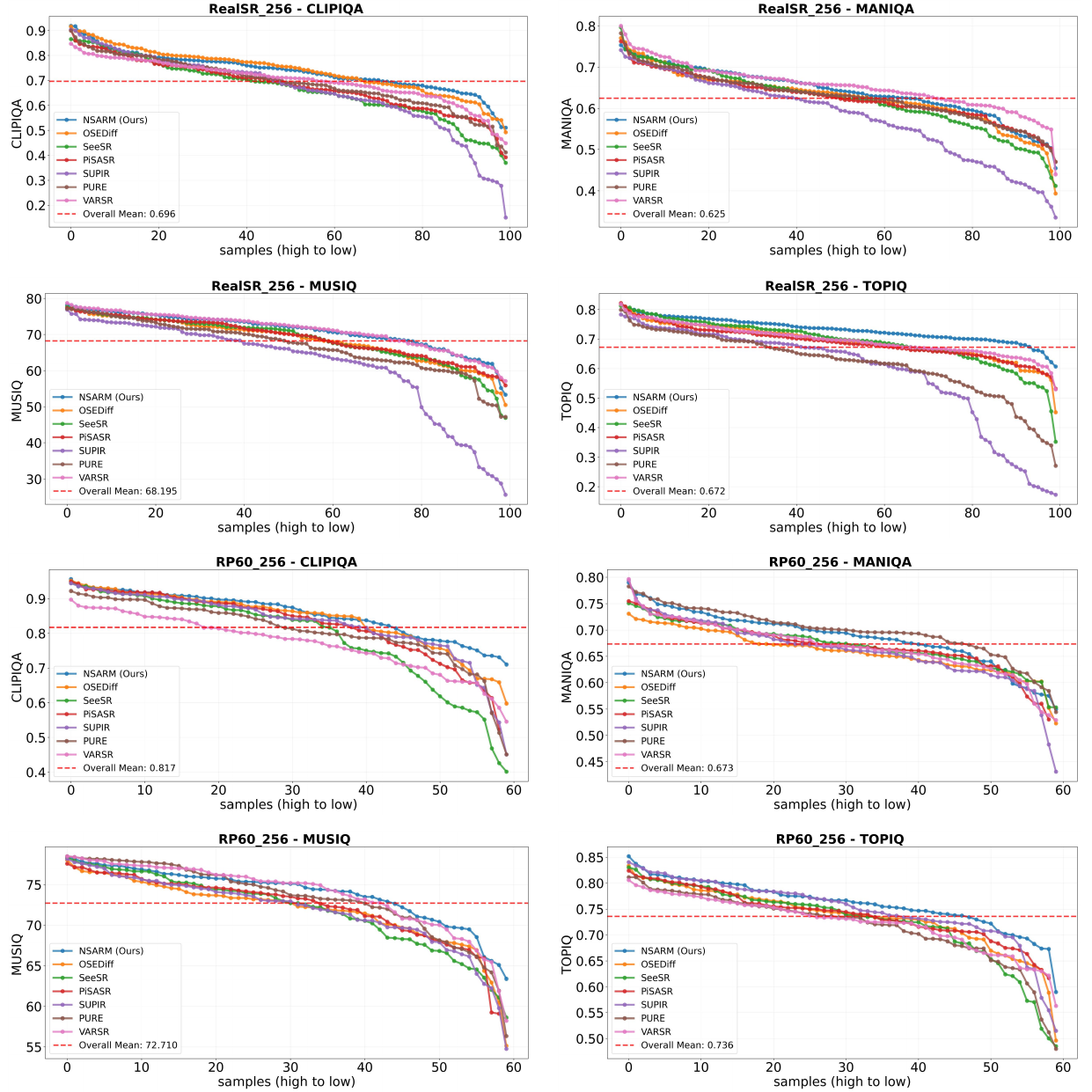}
\end{center}
\caption{Detailed ranked score distributions of Real-ISR methods across multiple evaluation metrics on RealSR and RP60.}
\label{fig:dis2}
\end{figure*}

\vspace{+1mm}
\noindent\textbf{More Sorted Score Distributions.}
In Fig.~\ref{fig:robust} of the main paper, we presented three representative curves showing sorted average scores across CLIP-IQA, MUSIQ, MANIQA (normalized by 100) and TOPIQ metrics for different methods on RealSR, RP60, and DIV2K datasets. For a more comprehensive analysis, we provide the per-metric and per-dataset results in Figs.~\ref {fig:dis1} and \ref{fig:dis2} (16 curves in total). These extended results demonstrate that: (1) the curves of NSARM are higher than other approaches in most cases, maintaining superior positions throughout the score distributions; and (2) the gradual descent of NSARM's curves, which indicates fewer bad quality outliers, confirms the robustness of NSARM. This consistent performance further validates NSARM's superior reconstruction quality and exceptional robustness.

\begin{figure*}[t]
\begin{center}
\includegraphics[width=1\linewidth]{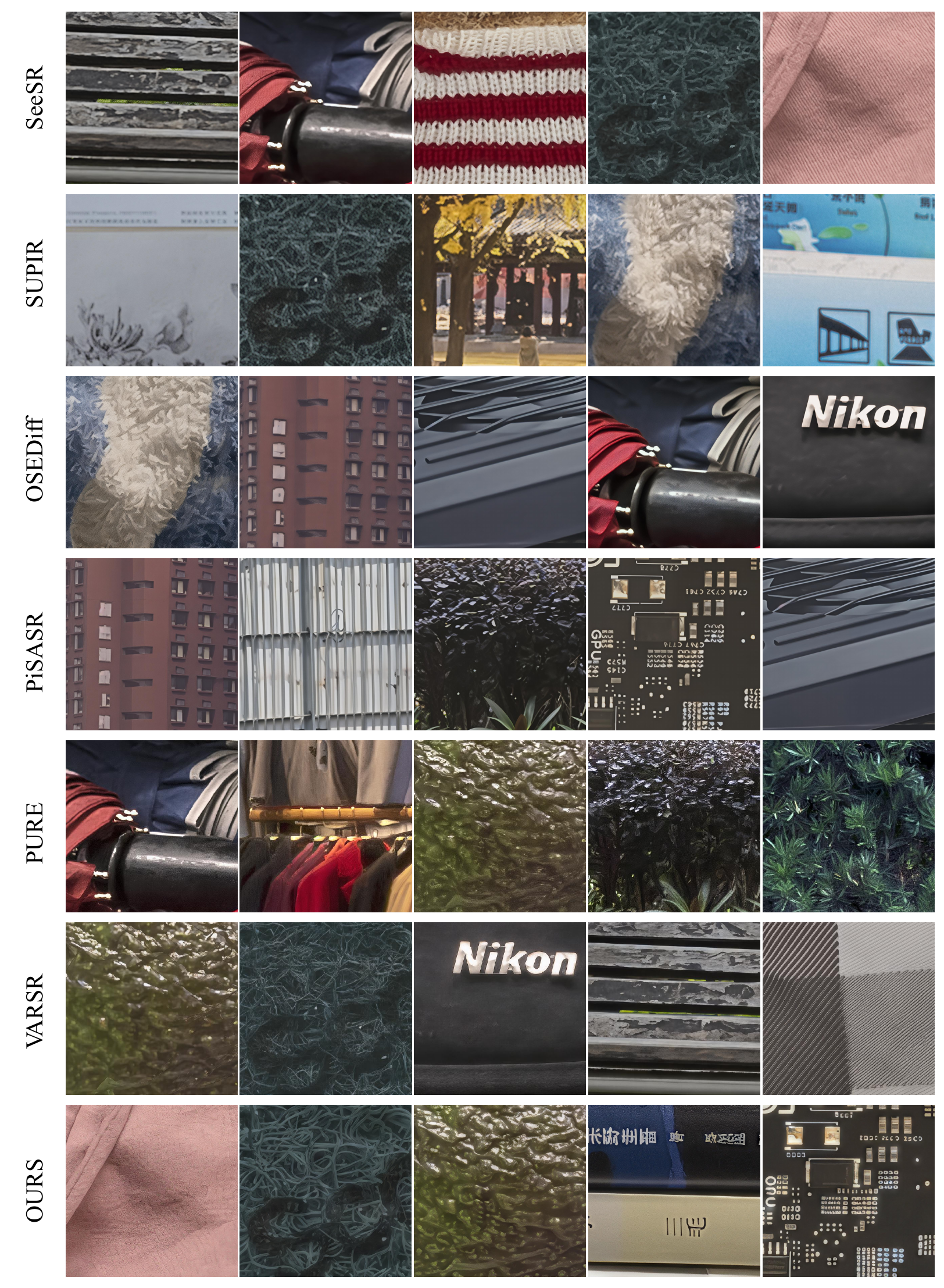}
\end{center}
\caption{The visual results of different methods with the worst perceptual metrics on RealSR testset.}
\label{fig:worst_realsr}
\end{figure*}

\begin{figure*}[t]
\begin{center}
\includegraphics[width=1\linewidth]{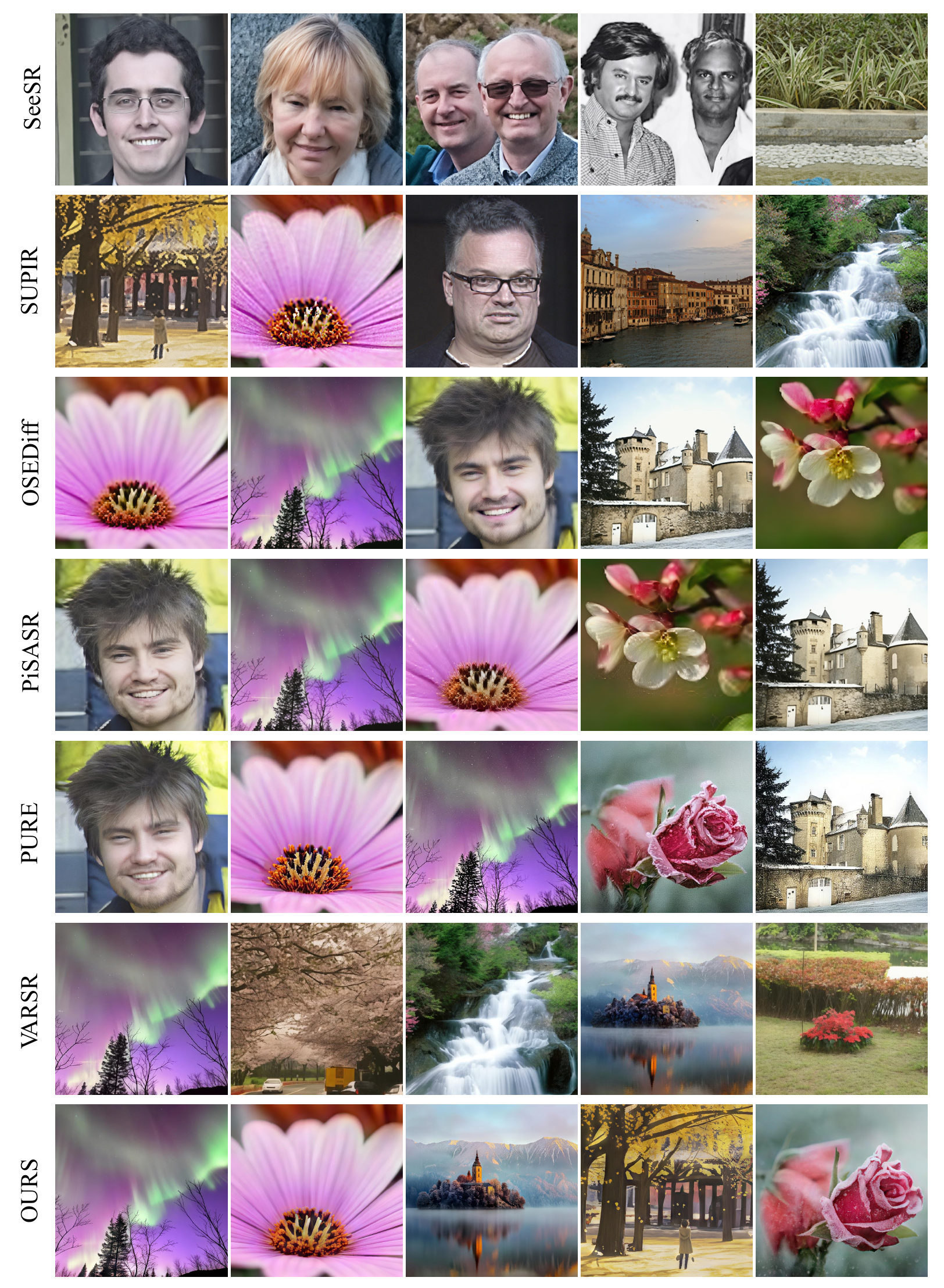}
\end{center}
\caption{The visual results of different methods with the worst perceptual metrics on RP60 testset.}
\label{fig:worst_RP}
\end{figure*}

\begin{figure*}[t]
\begin{center}
\includegraphics[width=1\linewidth]{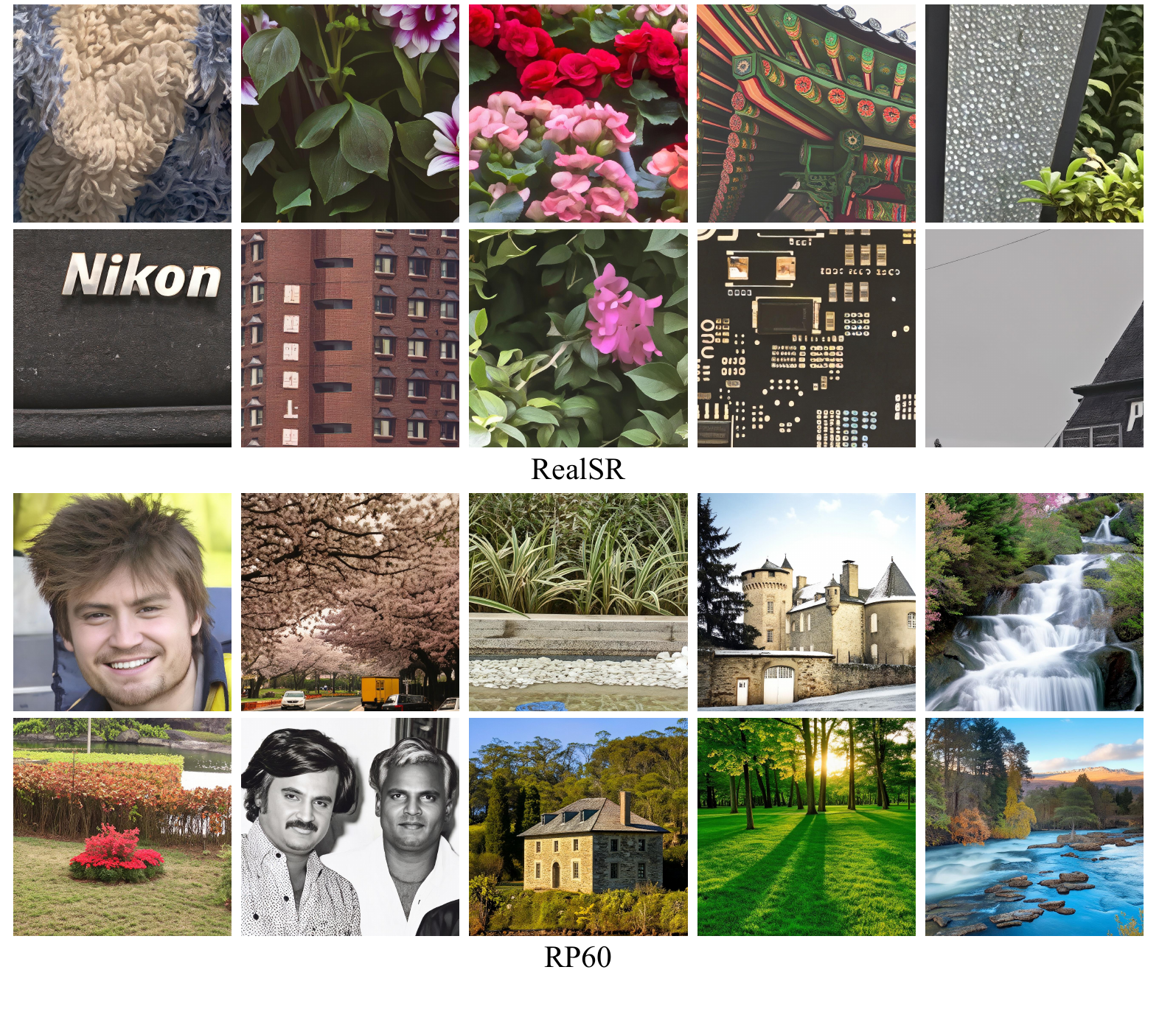}
\end{center}
\caption{The images that NSARM surpasses the others the most in terms of perception metrics.}
\label{fig:best}
\end{figure*}

\vspace{+1mm}
\noindent\textbf{Discussions on Worst Cases.}
In the curves presented in Fig.~\ref{fig:robust} of the main paper and Figs.~\ref {fig:dis1} and \ref{fig:dis2} of the supplementary material, we rank the scores of various metrics of each method to evaluate the model robustness. Although these curves visualize the distribution of metric scores across different samples, the comparison at each position of the abscissa may not correspond to the same input image. Therefore, in this section, we show in  Figs.~\ref {fig:worst_realsr} and \ref{fig:worst_RP} the five images with the lowest perceptual metric scores (the average scores over CLIPIQA, MUSIQ, MANIQA divided by 100, and TOPIQ) for each method on the RealSR and RP60 datasets, respectively, revealing the specific images that receive poor scores. We can see that the images with the worst scores largely overlap among methods, indicating that our approach genuinely improves the performance on challenging images where many methods fail. In Fig.~\ref{fig:best}, we show the images that NSARM performs better than others, sorted by the improvement between NSARM's scores and the average scores over competing methods. Notably, many of these images coincide with the worst-case images shown in Figs.~\ref {fig:worst_realsr} and \ref{fig:worst_RP}. This consistency further validates that our method achieves better performance on those low-scoring images, demonstrating its robustness and generalization performance.

\begin{table*}[tbp]
    \centering

    \adjustbox{width=\textwidth,center}{
\begin{tabular}{l|l|cccccccccccc}
    \toprule
    \multirow{2}{*}{Datasets} & \multirow{2}{*}{Methods} & \multirow{2}{*}{PSNR↑} & \multirow{2}{*}{SSIM↑} & \multirow{2}{*}{LPIPS↓} & \multirow{2}{*}{NIQE↓} & \multicolumn{2}{c}{MANIQA} & \multicolumn{2}{c}{MUSIQ} & \multicolumn{2}{c}{CLIPIQA}  & \multicolumn{2}{c}{TOPIQ}\\
     & & & & & & Avg.↑ & Var.↓ & Avg.↑ & Var.↓ & Avg.↑ & Var.↓ & Avg.↑ & Var.↓ \\
    \midrule
    
    \multirow{7}{*}{DRealSR} 
        & SeeSR~\cite{wu2024seesr} & 28.76 & 0.7762 & 0.3316 & 7.51 & 0.5596 & 0.0124 & 61.11 & 174.4791 & 0.6401 & 0.0217 & 0.6217 & 0.0207 \\
     & SeeSR (512) & 28.07 & 0.7684 & 0.3174 & 6.41 & 0.6042 & 0.0068 & 65.09 & 111.4403 & 0.6908 & 0.0152 & 0.6575 & 0.0111 \\
     & OSEDiff~\cite{wu2024one} & 28.60 & 0.7872 & 0.3296 & 7.94 & 0.5373 & 0.0087 & 60.91 & 142.5503 & 0.7082 & 0.0161 & 0.6208 & 0.0113 \\
    & OSEDiff (512) & 27.92 &0.7836 & 0.2967 & 6.44 & 0.5895 & \textcolor{red}{0.0038} & 64.68 & \textcolor{red}{86.7438} & 0.6952 & 0.0136 & 0.6002 & \textcolor{red}{0.0051} \\
    & PiSA-SR~\cite{sun2025pixel} & \textcolor{red}{28.88} & \textcolor{red}{0.7974} & \textcolor{red}{0.3145} & 7.81 & 0.5243 & 0.0091 & 60.33 & 131.9326 & 0.6453 & 0.0217 & 0.5987 & 0.0130 \\
    & PiSA-SR (512) & 28.32 & 0.7804 & 0.2960 &\textcolor{red}{ 6.18} & \textcolor{red}{0.6153} & 0.0046 & 66.11 & 92.9261 & 0.6967 & 0.0123 & 0.6334 & 0.0060 \\
 & NSARM (Ours) & 26.66 & 0.6914 & 0.4038 & 7.56 & 0.5903 & 0.0069 & \textcolor{red}{67.67} & 89.8052 & \textcolor{red}{0.7259} & \textcolor{red}{0.0103} & \textcolor{red}{0.6991} & 0.0054 \\
    \midrule
         \multirow{7}{*}{RealSR} 
    & SeeSR~\cite{wu2024seesr} & 25.52 & 0.7369 & 0.3065 & 6.79 & 0.6202 & 0.0056 & 68.57 & 44.2618 & 0.6663 & 0.0141 & 0.6909 & 0.0063 \\
     & SeeSR (512) & 25.42 & 0.7295 & 0.2984 & \textcolor{red}{5.39} & 0.6396 & 0.0041 & 69.25 & 37.6640 & 0.6537 & 0.0134 & 0.6795 & 0.0076 \\
     & OSEDiff~\cite{wu2024one} & 25.52 & 0.7334 & 0.3275 & 6.66 & 0.6267 & 0.0043 & 68.65 & 34.3655 & \textcolor{red}{0.7393} & 0.0081 & 0.6924 & 0.0035 \\
    & OSEDiff (512) & 25.15 & 0.7341 & 0.2920 & 5.64 & 0.6330 & \textcolor{red}{0.0023} & 69.10 & 25.9077 & 0.6690 & 0.0126 & 0.6253 & 0.0039 \\
    & PiSA-SR~\cite{sun2025pixel} & \textcolor{red}{25.91} & \textcolor{red}{0.7526} & \textcolor{red}{0.2902} & 6.35 & 0.6254 & 0.0031 & 69.14 & 29.6005 & 0.6803 & 0.0104 & 0.6885 & 0.0029 \\
    & PiSA-SR (512) & 25.50 & 0.7418 & \textcolor{red}{0.2672} & 5.51 & \textcolor{red}{0.6552} & 0.0024 & 70.15 & 26.3891 & 0.6699 & 0.0111 & 0.6373 & 0.0044 \\
 & NSARM (Ours) & 23.52 & 0.6399 & 0.3706 & 6.81 & 0.6402 & 0.0037 & \textcolor{red}{71.09} & \textcolor{red}{24.5262} & 0.7370 & \textcolor{red}{0.0061} & \textcolor{red}{0.7318} & \textcolor{red}{0.0016} \\
    \midrule
         \multirow{7}{*}{RP60} 
    & SeeSR~\cite{wu2024seesr} & - & - & - & 5.16 & 0.6717 & 0.0018 & 71.86 & 21.8666 & 0.7881 & 0.0184 & 0.7253 & 0.0061 \\
     & SeeSR (512) & - & - & - & 4.03 & 0.6649 & 0.0020 & 72.20 & 23.4764 & 0.7654 & 0.0128 & 0.7268 & 0.0056 \\
  & OSEDiff~\cite{wu2024one} & - & -& -& 5.49 & 0.6589 & \textcolor{red}{0.0016} & 71.84 & 16.9560 & 0.8413 & 0.0069 & 0.7351 & 0.0036 \\
      & OSEDiff (512) & - & - & - & 3.88 & 0.6551 & 0.0017 & 71.29 & 22.1331 & 0.7480 & 0.0073 & 0.6525 & 0.0042 \\
       & PiSA-SR~\cite{sun2025pixel} & - & - & - & 5.70 & 0.6597 & 0.0021 & 71.18 & 15.7543 & 0.8145 & 0.0090 & 0.7282 & 0.0023 \\
      & PiSA-SR (512)& - & - & - & \textcolor{red}{3.64} & \textcolor{red}{0.6889} & 0.0018 & 73.50 & 13.6328 & 0.7800 & 0.0072 & 0.7068 & 0.0029 \\
      & NSARM (Ours) & - &-& - & 5.69 &0.6862 & 0.0027 & \textcolor{red}{73.95} & \textcolor{red}{11.5072} & \textcolor{red}{0.8564} & \textcolor{red}{0.0040} & \textcolor{red}{0.7615} & \textcolor{red}{0.0021} \\
    \bottomrule
    \end{tabular}
    }
    \caption{Performance comparison of different Real-ISR methods on commonly used datasets. ``(512)" means that the method directly generates $512 \times 512$ results. \textcolor{red}{Red} color represents the best performance. ↓ and  ↑ represent the smaller or bigger is better.}
    \label{tab:methods_comparison_512}
\end{table*}

\end{document}